\documentclass[12pt]{article}

\usepackage{times,epsfig,graphics,amssymb,latexsym,amsmath,setspace,amsthm, fullpage, subcaption}
\usepackage{array,hyperref,url}
\usepackage[usenames]{color}
\usepackage[ruled, vlined]{algorithm2e}

\newtheorem{theorem}{\bf Theorem}
\newtheorem{Corollary}{\bf Corollary}

\def\argmin{\mathop{\rm argmin}}

\def\Var{\mathop{\rm Var}}

\def\E{\mathop{\rm E}}

\newcommand {\bfbeta} {\mbox{\boldmath $\beta$}}
\newcommand {\bfalpha} {\mbox{\boldmath $\alpha$}}

\newcommand {\bftheta} {\mbox{\boldmath $\theta$}}
\newcommand {\bfgamma} {\mbox{\boldmath $\gamma$}}

\def\bv{\mathop{\bf v}}

\def\bc{\mathop{\bf c}}

\def\bx{\mathop{\bf x}}
\def\bX{\mathop{\bf X}}

\def\bK{\mathop{\bf K}}

\def\by{\mathop{\bf y}}

\def\bI{\mathop{\bf I}}

\def\bu{\mathop{\bf u}}
\def\bv{\mathop{\bf v}}

\def\bu{\mathop{\bf u}}

\begin{document}

\pagenumbering{arabic}

\setcounter{page}{1}

\title{\Large \bf Efficient kernel-based variable selection with sparsistency   }
\author{
     Xin He$^\dag$, Junhui Wang$^{\ddag}$ and Shaogao Lv$^{\S}$\\[10pt]
     $^\dag$ School of Statistics and Management\\
     Shanghai University of Finance and Economics \\
     \and
     $^\ddag$ School of Data Science \\
          City University of Hong Kong \\
     \and
     $^{\S}$ School of Statistics and Mathematics \\
    	Nanjing Audit University
}
\date{}

\maketitle

\doublespacing
\begin{abstract}
	Sparse learning is central to high-dimensional data analysis, and various methods have been developed. Ideally, a sparse learning method shall be methodologically flexible, computationally efficient, and with theoretical guarantee, yet most existing methods need to compromise some of these properties to attain the other ones. In this article, a three-step sparse learning method is developed, involving kernel-based estimation of the regression function and its gradient functions as well as a hard thresholding. Its key advantage is that it assumes no explicit model assumption, admits general predictor effects, allows for efficient computation, and attains desirable asymptotic sparsistency. The proposed method can be adapted to any reproducing kernel Hilbert space (RKHS) with different kernel functions, and its computational cost is only linear in the data dimension. The asymptotic sparsistency of the proposed method is established for general RKHS under mild conditions. Numerical experiments also support that the proposed method compares favorably against its competitors in both simulated and real examples.
\end{abstract}

\section{Introduction}
sparse learning has attracted tremendous interests from both researchers and practitioners, due to the availability of large number of variables in many real applications. In such scenarios, identifying the truly informative variables for the objective of analysis has become a key factor to facilitate statistical modeling and analysis. Ideally, a sparse learning method shall be flexible, efficient, and with theoretical guarantee. To be more specific, the method shall not assume restrictive model assumptions, so that it is applicable to data with complex structures; its implementation shall be computationally efficient and able to take advantage of high performance computing platform; it shall have theoretical guarantee on its asymptotic consistency in identifying the truly informative variables. 

In literature, many sparse learning methods have been developed in the regularization framework assuming certain working model set. The most popular working model set is to assume a linear model, where the sparse learning task simplifies to identifying nonzero coefficients. Under the linear model assumption, the regularization framework consists of a least square loss function for the linear model as well as a sparsity-inducing regularization term. Various regularization terms have been considered, including the least absolute shrinkage and selection operator (Lasso; \cite{Tibshirani1996}), the smoothly clipped absolute deviation (SCAD; \cite{Fan2001}), the adaptive Lasso \cite{ZouH2006}, the minimax concave penalty (MCP; \cite{ZhangC2010}), the truncated $l_1$-penalty (TLP; \cite{Shen2012}), {the $l_0$-penalty \cite{Shen2013}}, and so on. These methods have also been extended to the nonparametric models to relax the linear model assumption. For example, under the additive model assumption, a number of sparse learning methods have been developed   \cite{Shively1999, Huang2010}, where each component function depends on one variable only. {Further, a component selection and smoothing operator method (COSSO; \cite{Lin2006}) is proposed to allow higher-order interaction components in the additive model. {Yet the higher-order additive models need to enumerate all the interaction components, which may explode in an exponential order of the number of variables.} These nonparametric sparse learning methods, although more flexible than the linear model, still require some explicit working model sets. }

More recently, attempts have been made to develop {nonparametric} sparse learning methods to circumvent the dependency on restrictive model assumptions. Particularly, sparse learning is formulated in a dimension reduction framework in Li et al. \cite{LiB2005} and Bondell and Li \cite{Bondell2009} via searching for the sparse basis of the central dimension reduction space. Fukumizu and Leng \cite{Fukumiza2014} developed a gradient-based dimension reduction method that can be extended to nonparametric sparse learning. A novel measurement-error-model-based sparse learning method is developed in Stefanski et al. \cite{Stafabski2014} and Wu and Stefanski \cite{WuY2015} for nonparametric kernel regression models, {and some gradient learning methods  \cite{Rosasco2013, YangL2016}  are proposed to conduct sparse learning in a flexible RKHS \cite{Wahba1998}.} Also, a flexible knock-off filter framework \cite{Barber2015} and a recursive feature elimination method by using kernel ridge regression are proposed \cite{Dasgupta2018}, which show substantial advantage than most existing methods, yet their lack of selection consistency or computational efficiency remain as some of their main obstacles. {Specifically, it is also interesting to point out that most existing gradient-based methods \cite{Rosasco2013, YangL2016} aim to directly  estimate the gradient functions in a regularization framework with some well-designed penalty terms,
and thus they may not be applicable to analyze data with high dimension due to their expensive computational cost. }

Another popular line of research on high-dimensional data is variable screening, which screens out uninformative variables by examining the marginal relationship between the response and each variable. The marginal relationship can be measured by various criteria, including the Pearson's correlation \cite{FAN2008}, the empirical functional norm \cite{Fan2011}, the distance correlation \cite{Szekely2007}, and a quantile-adaptive procedure \cite{Hexm2013}. All these methods are computationally very efficient, and attain the sure screening property, meaning that all the truly informative variables are retained after screening with probability tending to one. This is a desirable property, yet slightly weaker than the asymptotic consistency in sparse learning. Another potential weakness of the marginal screening methods is that they may ignore those marginally unimportant but jointly important variables \cite{Hexm2013}. To remedy this limitation, some recent work \cite{Hao2014, WangX2016} has been done to conduct sure screening for variables with interaction effects.  

In this article, we propose an efficient kernel-based sparse learning method, which is methodologically flexible, computationally efficient, and able to achieve the asymptotic consistency without requiring any explicit model assumption.  
The method consists of three simple steps, involving kernel-based estimation of the regression function and its gradient functions as well as a hard thresholding. It first fits a kernel ridge regression model in a flexible RKHS to obtain an estimated regression function, then estimates its gradient functions along each variable by taking advantage of the derivative reproducing property \cite{ZhouD2007}, and finally hard-thresholds the empirical norm of each gradient function to identify the truly informative variables. This method is flexible in that it can be adapted to any RKHS with different kernel functions, to accommodate prior information about the true regression function. The proposed method also enables efficient estimation of the gradient functions in two steps   by using the derivative property in RKHS, which  significantly reduces the computational cost and allows for  diverging dimension. Its computational cost is only linear in the data dimension, and thus computationally efficient  to analyze dataset with large dimensions. For example, the simulated examples with $p=100000$ variables can be efficiently analyzed on a standard multi-core PC. More importantly, asymptotic consistency can be established for the proposed method without requiring any explicit model assumptions. It is clear that the proposed method is advantageous than the existing methods, as it achieves methodological flexibility, numerical efficiency and asymptotic consistency. To our knowledge, this method is the first one that can achieve these three desirable properties at the same time.

The rest of the article is organized as follows. In Section \ref{Sec2}, we present the proposed general kernel-based sparse learning method as well as its computational scheme. In Section \ref{Sec3}, the asymptotic consistency of the proposed method is established. {Two theoretical examples are provided in In Section \ref{theoreticalexample:1}.} In Section \ref{Sec4}, the proposed method is extended to select truly informative interaction terms. The numerical experiments on the simulated and real examples are contained in Section \ref{Sec5}, followed by a concluding summary in Section \ref{Sec6}. All the necessary lemmas and technical proofs are given in Appendix and supplementary materials.


\section{Proposed method}\label{Sec2}

\subsection{Regression in RKHS}

Suppose a random sample ${\cal Z}^n=\{(\bx_i,y_i)\}_{i=1}^n$ are independent copies of ${\cal Z}=(\bx,y)$, drawn from some unknown distribution $\rho_{\bx,y}$ with ${\bx}=(x^{1} , ..., x^p)^T\in {\cal X}$ supported on a compact metric space and $y \in {\cal R}$. Consider a general regression setting,
$$
y=f^*(\bx)+\epsilon,
$$
where $\epsilon$ is a random error with $\E(\epsilon|\bx)=0$ and $\Var(\epsilon|\bx)=\sigma^2$, and thus $f^*(\bx)=\int yd\rho_{y|\bx}$ with $\rho_{y|\bx}$ denoting the conditional distribution of $y$ given $\bx$. It is also assumed that $f^* \in {\cal H}_K$, where ${\cal H}_{K}$ is a RKHS induced by some pre-specified kernel function $K(\cdot,\cdot)$. For each ${\bx} \in {\cal X}$, denote $K_{\bx}=K(\bx,\cdot)\in {\cal H}_{K}$, and the reproducing property of RKHS implies that $\langle f, {K}_{\bx} \rangle_{K}=f(\bx)$ for any $f \in {\cal H}_K$, where $\langle \cdot, \cdot \rangle_K$ is the inner product in ${\cal H}_{K}$. 

The RKHS enjoys a number of desirable properties making it particularly suitable for general nonparametric models, including its approximation ability, its functional complexity and derivative reproducing property. To be precise, many popularly used kernels, including Gaussian kernel and Laplace kernel, are universal \cite{Steinwart2008}, meaning that their induced RKHS's are dense in the continuous function space under the infinity norm. This  universal approximation property ensures that the kernel-based methods can yield nonparametric estimates with small approximation error in estimating any continuous target function. On the other hand, to characterize statistical properties for nonparametric models, the notion of functional complexity appearing in empirical process are widely employed for theoretical analysis, such as various covering numbers, VC dimension and Rademacher complexity \cite{Bartlett2002}. The RKHS has a very interesting and surprising property that for a unit ball $B_1$ of the RKHS, its Rademacher complexity \cite{Bartlett2002} can be bounded as $R_n(B_1)\leq 2n^{-1/2}(\mathbb{E}(K(X,X)))^{1/2}$, where $R_n(\cdot)$ denotes the global Rademacher complexity. In other words, the functional complexity of the bounded ball in the RKHS is less affected by the dimension of variables, and thus a small variance estimator without sacrificing approximation ability for nonparametric estimation can be obtained by kernel-based methods. In addition, in the literature of nonparametric statistics, estimating the gradient function of the target function is generally hard. However, the derivative of any function in a smooth RKHS also has the reproducing property, implying that kernel-based methods have simultaneous convergence behavior in both the function itself and its gradient function with the same rate of convergence under the sup norm. 

\subsection{Gradient-based sparse learning}

In sparse modeling, it is generally believed that $f^*(\bx)$ only depends on a small number of variables, while others are uninformative. Unlike model-based settings, sparse learning for a general regression model is challenging due to the lack of explicit regression parameters. Here we measure the importance of variables in a regression function by examining the corresponding gradient functions. It is crucial to observe that if a variable $x^l$ is deemed uninformative, the corresponding gradient function 
$$
g^*_l(\bx)={\partial f^*(\bx)}/{\partial {x}^l }
$$ 
should be exactly zero almost surely. Thus the true active set can be defined as 
$$
{\cal A}^*=\{l: \left \| g^*_l \right \|^2_{2} >0 \},
$$
where $ \| g^*_l  \|^2_{2} = \int \left ( g^*_l(\bx) \right )^2 d\rho_{\bx}$ with the marginal distribution $\rho_{\bx}$. 

The proposed general sparse learning method is presented in Algorithm \ref{alg:general_VS}.
\begin{figure}[!ht]
	\centering
	\begin{minipage}{.9\linewidth}	
	\begin{algorithm}[H]
			\label{alg:general_VS}
			\SetAlgoLined
			Step 1: Obtain an estimate $\widehat{f}$ in a smooth RKHS based on the given sample ${\cal Z}^n$; \\
			Step 2: Compute $\widehat{g}_l(\bx) = \partial \widehat{f}(\bx) / \partial x^l $ for $l=1,\ldots,p$; \\
			Step 3: Identify the informative variables by checking the norm of each $\widehat{g}_l$.
			\caption{General sparse learning method}
		\end{algorithm}
	\end{minipage}
\end{figure}
	\\
	
We now give details of each step in Algorithm \ref{alg:general_VS}. To obtain $\widehat{f}$ in Step 1, we employ the kernel ridge regression model, 
\begin{align}\label{eqn:e1}
\widehat f(\bx) = \argmin_{f \in {\cal H}_{K}} \ \frac{1}{n}\sum_{i=1}^n \left (y_i- f({\bx}_i) \right )^2   + \lambda_n \|f\|^2_{K},
\end{align}
where the first term, denoted as ${\cal E}_n (f)$, is an empirical version of ${\cal E}(f)=\E(y-f(\bx))^2$, and $\|f\|_{K}=\langle f, f \rangle_K^{1/2}$ is the associated RKHS-norm of $f \in {\cal H}_K$. By the representer theorem \cite{Wahba1998}, the minimizer of (\ref{eqn:e1}) must have the form
$$
\widehat{f}(\bx)=\sum_{i=1}^n\widehat{\alpha}_i K({\bx}_i,\bx) = \widehat{\bfalpha}^T {\bK}_{n}(\bx),
$$
where $\widehat \bfalpha=(\widehat \alpha_1,...,\widehat \alpha_n)^T$ and ${\bK}_{n}(\bx)=(K({\bx}_1,\bx),...,K({\bx}_n,\bx))^T$.  
Then the optimization task in (\ref{eqn:e1}) can be solved analytically, with
\begin{align} \label{eqn:alpha}
\widehat{\bfalpha}= \left({\bK}^2 + n \lambda_n \bK \right)^{+}  {\bK}\by,
\end{align}  
where $\bK=\big( K({\bx}_i,{\bx}_j) \big )_{i,j=1}^n$, and ${}^+$ denotes the Moore-Penrose generalized inverse of a matrix. When $\bK$ is invertible, \eqref{eqn:alpha} simplifies to $\widehat{\bfalpha}= \left({\bK} + n \lambda_n \bI \right)^{-1} \by$. 

Next, to obtain $\widehat g_l$ in Step 2, it follows from Lemma 1 in  the supplementary material that for any $f \in {\cal H}_K$,
$$
g_l(\bx)=\frac{\partial f(\bx)}{\partial x^l} =\langle f, {\partial_l {K}_{\bx}} \rangle_K \leq \|\partial_{l}K_{\bx}\|_K\| f\|_K,
$$ 
where $ {\partial_l {K}_{\bx}}=\frac{\partial K(\bx,\cdot)}{\partial x^l}$. This implies that the gradient function of any $f \in {\cal H}_{K}$ can be bounded by its $K$-norm up to some constant. In other words, if we want to estimate $g_l^*(\bx)$ within the smooth RKHS, it suffices to estimate $f^*$ itself without loss of information. Consequently, if $\widehat{f}$ is obtained in Step 1, $g_l^*(\bx)$ can be estimated as $\widehat{g}_l(\bx)=\widehat{\bfalpha}^T {\partial_l {\bK}_n({\bx}})$ for each $l$,  where ${\partial_l {\bK}_n({\bx})}=({\partial_l {K}_{\bx}}({\bx}_1),...,{\partial_l {K}_{\bx}}({\bx}_n))^T$.

In Step 3, it is difficult to evaluate $\| \widehat{g}_l \|^2_{2}$ directly, as $\rho_{\bx}$ is usually unknown in practice. We then adopt the empirical norm of $\widehat{g}_l$ as a practical measure, 
\begin{align*}
\| \widehat{g}_l \|^2_n= \frac{1}{n}\sum_{i=1}^n \big( \widehat{g}_l({\bx}_i) \big )^2= \frac{1}{n} \sum_{i=1}^n \big ( \widehat{\bfalpha}^T {\partial_l {\bK}_n({\bx}_i)}  \big )^2.
\end{align*}
The estimated active set can be set as {$\widehat{\cal A}_{v_n}= \left \{l: \left \|\widehat{g}_l\right \|^2_n > v_n \right \}$} for some pre-specified $v_n$. It is clear that our method can be regarded as a nonparametric joint screening method, which can correctly identify  all the truly informative variables  acting on the response with a general effect, including those marginally noninformative but jointly informative ones.

The proposed method presented in Algorithm \ref{alg:general_VS} is general in that it can be adapted to any smooth RKHS with different kernel functions, where the choice of kernel function depends on prior knowledge about $f^*$. For instance, if $f^*$ is known as linear or polynomial function in advance, the RKHS induced by the linear or polynomial kernel can be used. If no prior information about $f^*$ is available, the RKHS induced by the Gaussian kernel can be used, which is known to be universal in the sense that any continuous function can be well approximated by some function in the induced RKHS under the infinity norm \cite{Steinwart2008}.  In practice, unless some reliable prior information about $f^*$ is known, it is recommended to consider the RKHS induced by the Gaussian kernel due to its capacity and flexibility. 

{{\bf Remark 1:} The proposed method is  computationally efficient, whose computational cost is about $O(n^3 + n^2p)$. The complexity $O(n^3)$ comes from inverting an $n \times n$ matrix in (\ref{eqn:alpha}), and the complexity $O(n^2p)$ comes from calculating $\| \widehat{g}_l \|^2_n$ for $l=1,\ldots,p$. This complexity is particularly attractive in the large-$p$-small-$n$ scenario, where the computational complexity becomes linear in $p$ and parallelization can be employed to further speed up the computation. In some other scenarios with large $n$, the $O(n^3)$ complexity can be too demanding. Some possible improvements are available to alleviate the computational burden by some low rank approximation, such as the random sketch method in Yang et al. \cite{YangY2017}. Its computational complexity can be reduced to $O(m^3)$, where $m \ll n$ is the sketch dimension to be determined as in \cite{YangY2017}. More importantly, the random sketch method is proved to be fast and minimax optimal for fitting the kernel ridge regression.

{\bf Remark 2:} The estimated regression function $\widehat{f}$ is merely an intermediate step for estimating the gradient functions, which is a consistent estimate but converges to the true regression function $f^*$ at some rather slow rate due to the inclusion of the noise variables.  We also want to emphasize that the data is only used once to estimate the representer coefficients $\widehat{\bfalpha}$ in \eqref{eqn:alpha}, and then the estimated gradient function $\widehat{g}_l$ can be estimated directly by using the derivative reproducing property in RKHS by Lemma 1 in  the supplementary material.

\subsection{Tuning}\label{Tune}

The proposed method presented in Algorithm \ref{alg:general_VS} consists of two tuning parameters, the ridge parameter $\lambda_n$ and the thresholding parameter $v_n$. Based on our limited numerical experience, the proposed method performs well and stable when the ridge parameter $\lambda_n$ is sufficiently small in various scenarios. Similar observation on the choice of ridge parameter has also been made in \cite{WangX2016}. Therefore, we set $\lambda_n=0.001$ and focus on the choice of $v_n$ in all the simulated experiments.

To optimize the selection performance of the proposed method, we employ the stability-based criterion \cite{SunWW2013} to select the value of $v_n$. Its key idea is to measure the stability of sparse learning by randomly splitting the training sample into two parts and comparing the disagreement between the two estimated active sets. Specifically, given a thresholding value $v_n$, we randomly split the training sample ${\cal Z}^n$ into two parts ${\cal Z}^n_1$ and ${\cal Z}^n_2$. Then the proposed method is applied to ${\cal Z}^n_1$ and ${\cal Z}^n_2$ and obtains two estimated active sets  $\widehat{\cal A}_{1,v_n}$ and $\widehat{\cal A}_{2,v_n}$, respectively. The disagreement between $\widehat{\cal A}_{1,v_n}$ and $\widehat{\cal A}_{2,v_n}$ is measured by Cohen's kappa coefficient
$$
\kappa(\widehat{\cal A}_{1,v_n},\widehat{\cal A}_{2,v_n})=\frac{Pr(a)-Pr(e)}{1-Pr(e)},
$$
where $Pr(a)=\frac{n_{11}+n_{22}}{p}$ and  $Pr(e)=\frac{(n_{11}+n_{12})(n_{11}+n_{21})}{p^2}+\frac{(n_{12}+n_{22})(n_{21}+n_{22})}{p^2}$ with $n_{11}=|\widehat{\cal A}_{1,v_n} \cap\widehat{\cal A}_{2,v_n}|, n_{12}=|\widehat{\cal A}_{1,v_n} \cap \widehat{\cal A}^C_{2,v_n}|,n_{21}=|\widehat{\cal A}^C_{1,v_n} \cap \widehat{\cal A}_{2,v_n}|, n_{22}=|\widehat{\cal A}^C_{1,v_n} \cap \widehat{\cal A}^C_{2,v_n}|$ and $|\cdot|$ denotes the set cardinality. 

The procedure is repeated for $B$ times and the estimated sparse learning stability is measured as
$$
\hat s(\Psi_{v_n})=\frac{1}{B}\sum_{b=1}^B \kappa(\widehat{\cal A}_{1,v_n}^b, \widehat{\cal A}_{2,v_n}^b).
$$
Finally, the thresholding parameter $\widehat{v}_n$ is set as $
\widehat{v}_n = \max \big \{ v_n: \frac{\hat s(\Psi_{v_n})}{\max_{v_n}\hat s(\Psi_{v_n})} \geq q \big \}$, {where $q \in (0,1)$ is some given percentage. In all the simulated experiments, we set $q=0.95$ as suggested in \cite{SunWW2013}, and the performance of the resultant tuning criterion appears to be satisfactory.

\section{Asymptotic sparsistency}\label{Sec3}

Now we establish the asymptotic consistency of the proposed method. First, we introduce an integral operator 
$L_K: {\cal L}^2({\cal X},{\rho_{\bx}}) \rightarrow  {\cal L}^2({\cal X},{\rho_{\bx}})$, given by 
\begin{align*}
L_K(f)(\bx)=\int K(\bx,\bu)f(\bu)d\rho_{\bx}(\bu), 
\end{align*}
for any $f \in {\cal L}^2({\cal X},{\rho_{\bx}})=\{f: \int f^2(\bx)d\rho_{\bx}<\infty\}$.  Note that if the corresponding RKHS is separable, by the spectral theorem we have
$$
L_Kf=\sum_{j}\mu_j\langle f , e_j \rangle_2e_j,
$$
where $\{e_j\}$ is an orthonormal basis of ${\cal L}^2({\cal X}, \rho_{\bx})$, $\mu_j$ is the eigenvalue of the integral operator $L_K$, and $\langle \cdot, \cdot\rangle_2$ is the inner product in ${\cal L}^2({\cal X},\rho_{\bx})$.  By Mercer's theorem, under some regularity conditions, the eigen-expansion of the kernel function is
$
K(\cdot,\cdot)=\sum_{j\geq 1}\mu_j e_j(\cdot)e_j(\cdot).
$
Therefore, the RKHS-norm of any $f \in {\cal H}_K$ can be written as
\begin{align*}
\|f\|^2_K= \sum_{j\geq 1} \frac{ \langle f, e_j \rangle^2_2}{\mu_j},
\end{align*}
which implies the decay rate of $\mu_j$ fully characterizes the complexity of the RKHS, and is closely related with various entropy numbers \cite{Steinwart2008}.

We denote the cardinality of the true active set ${\cal A}^*$  as $|{\cal A}^*|=p_0$, and both $p_0$ and $p$ are allowed to diverge with $n$. The following technical assumptions are made.

{\noindent \bf Assumption 1}: Suppose that $f^*$ is in the range of the $r$-th power of $L_K$, denoted as $L_{K}^r$, for some positive constant $r\in (1/2,1]$.

{\noindent\bf Assumption 2}: There exist some constants $\kappa_1$ and $\kappa_2$ such that $\sup\limits_{\bx \in {\cal X}} \|K_{\bx}\|_K\leq \kappa_1$, and $\sup\limits_{\bx \in {\cal X}}\|\partial_lK_{\bx}\|_K\leq \kappa_2$, for any $l=1,...,p.$ 

{\noindent\bf Assumption 3}: The distribution of  $\epsilon$ has a $q$-exponential tail with some function $q(\cdot)$; that is, there exists some constant $c_1>0$ such that $P(|\epsilon|>t)\leq c_1\exp\{-q(t)\}$ for any $t>0$.

In Assumption 1, the operator $L_K$ on ${\cal L}^2({\cal X},{\rho_{\bx}})$ is self-adjoint and semi-positive definite, and thus its fractional operator $L_{K}^r$ is well defined. Furthermore,  the range of $L_K^r$ is contained in ${\cal H}_K$ if $r\geq 1/2$ \cite{Smale2007}, and thus Assumption 1 implies that there exists some function $h\in {\cal L}^2({\cal X},\rho_{\bx})$ such that $f^*=L^r_K h =\sum_{j}\mu^r_j\langle h , e_j \rangle_2e_j \in {\cal H}_K$, ensuring strong estimation consistency under the RKHS-norm. Similar assumptions are also imposed in  \cite{Mendelson2010}. {Assumption 2 assumes the boundedness of the kernel function and its gradient functions, and is satisfied by many popular kernels, including the Gaussian kernel and the Sobolev kernel \cite{Smale2007, Rosasco2013, YangL2016} with the compact support condition. Note that the compact support condition is commonly used in machine learning literature \cite{Mendelson2010, Rosasco2013, Dasgupta2018, Lv2018} for mathematical simplicity, and it may be  relaxed by allowing the support to expand with sample size, which leads to some additional treatment in the asymptotic analysis.} Assumption 3 characterizes the tail behaviour of the error distribution, which relaxes the commonly-used bounded in machine learning literature \cite{Smale2007, Rosasco2013, Lv2018}. It is general and satisfied by a variety of distributions \cite{WangX2016, ZhangC2016}. For example, if $\epsilon$ follows a sub-Gaussian distribution or any bounded distribution, Assumption 3 is satisfied with $q(t)=O(t^2)$; if $\epsilon$ follows a sub-exponential distribution, Assumption 3 is satisfied with $q(t)=O(\min\{\frac{t}{C}, \frac{t^2}{C^2} \})$ for some constant $C$. 


\begin{theorem}\label{thm1}
	Suppose Assumptions 1--3 are satisfied. Then with probability at least $1-\delta_n/2$, there holds
	\begin{align}\label{thm12}
	\big \|\widehat{f}-f^* \big \|_{K} \leq  2\log{\frac{8}{\delta_n}} \Big ( 3\kappa_1 \lambda_n^{-1}{n}^{-{1}/{2}} (\kappa_1\|f^*\|_K+q^{-1}(\log\frac{4c_1 n}{\delta_n})) + \lambda_n^{r-{1}/{2}} \|L^{-r}_{K}f^*\|_2 \Big ).
	\end{align}
	Additionally, let $\lambda_n={n^{-\frac{1}{2r+1}}}$, then with probability at least $1-\delta_n$, there holds
	\begin{align}\label{thm13}
	\max_{1 \leq l \leq p} \ \big |  \| \widehat{g}_l \|^2_n -  \| g^*_l \|_{2}^2 \big |  \leq b_{n, 1} \max \{ \kappa_1\|f^*\|_K, q^{-1}(\log\frac{4c_1 n}{\delta_n}) \}  {\log\Big(  \frac{8p}{\delta_n}\Big)} { n^{-\frac{2r-1}{2(2r+1)}}  } ,
	\end{align}
	where $b_{n, 1}=4\max\{\kappa^2_2, \kappa^2_2\|f^*\|_{K}, \|f^*\|^2_{K}\} \max\{3\kappa_1,  2\sqrt{2}\kappa_2^2, \|L_K^{-r}f^*\|_2\}$ and $q^{-1}(\cdot)$ denotes the inverse function of $q(\cdot)$.
\end{theorem}

Theorem \ref{thm1} establishes the convergence rate of the difference between the estimated regression function and the true regression function in terms of the RKHS-norm. 
Note that similar results  have been established in learning theory literature \cite{Smale2005, Smale2007}. Yet  these results assume  that the response to be uniformly bounded above, which can be too restrictive in practice.  Theorem \ref{thm1}  relaxes the restrictive boundness condition  by characterizing the tail behaviour of the error term. Theorem \ref{thm1} also shows that $\left \|\widehat{g}_l\right \|^2_n$ converges to $\left \| g_l^* \right \|_{2}^2$ with high probability, which is crucial to establish the asymptotic sparsistency. Note that $b_{n,1}$ is spelled out precisely for the subsequent analysis of the asymptotic sparsistency and its dependency on $f^*$. {Note that the convergence result still holds even when $p$ diverges with $n$, and the quantities $\|f^*\|^2_{K}$ and  $\|L^{-r}_{K}f^*\|_2$ in (\ref{thm12}) and (\ref{thm13}) may depend on $p_0$ through $f^*$, and thus may also diverges with $n$. Yet such dependencies are generally difficult to quantify explicitly in a fully general case \cite{Fukumiza2014}.}

{\bf Remark 3:} The rate of convergence in Theorem \ref{thm1} can be strengthened to obtain an optimal strong convergence rate in a minimax sense as in  \cite{Fischer2017}. Yet it requires the random error $\epsilon$ follows a sub-Gaussian distribution and the decay rate of $L_K$'s eigenvalues has an upper bound of polynomial order; that is, $\mu_j\leq C j^{-1/\tau}$ for some positive constant $C$ and $\tau\in(0,1)$. Then the rate of convergence in \eqref{thm13}  can be further improved. 



{\noindent\bf Assumption 4}: There exists some positive constant $\xi_1 < \frac{2r-1}{2(2r+1)}$ such that
$\min_{l \in {\cal A}^*}\left \| g^*_l \right \|^2_{2}> b_{n,1} \max \{ \kappa_1\|f^*\|_K, q^{-1}(\log\frac{4c_1 n}{\delta_n}) \} n^{-\xi_1} \log p.$
	
	Assumption 4 requires the true gradient function contains sufficient information about the truly informative variables. Unlike most nonparametric models, we measure the significance of each gradient function to distinguish the informative and uninformative variables without any explicit model specification. {Note that  the required minimal signal strength in Assumption 4 is much tighter than that in many nonparametric sparse learning methods \cite{Huang2010, YangL2016}, which often require the signal to be bounded away from zero.}
	
	Now we establish the asymptotic sparsistency of the proposed sparse learning method.
	
	\begin{theorem}\label{thm3}
		Suppose the assumptions of Theorem \ref{thm1} and Assumption 4 are satisfied. Let $v_n=\frac{b_{n,1}}{2} \max \{ \kappa_1\|f^*\|_K, q^{-1}(\log\frac{4c_1 n}{\delta_n}) \} n^{-\xi_1}\log p$, then we have
		\begin{align*}
		P\big( \widehat{\cal A}_{v_n}={\cal A}^* \big) \rightarrow 1, \ \ \mbox{as} \ \ n\rightarrow \infty.
		\end{align*}
	\end{theorem}
	
	Theorem \ref{thm3} shows that the selected active set can exactly recover the true active set with probability tending to 1. This result is particularly interesting given the fact that it is established for any RKHS with different kernel functions. A direct application of the proposed method and Theorem \ref{thm3} is to conduct nonparametric sparse learning with sparsistency {\cite{Szekely2007, Hexm2013,  YangL2016}}. If no prior knowledge about the true regression function is available, the proposed method can be applied with a RKHS associated with the Gaussian kernel. Asymptotic sparsistency can be established following Theorem \ref{thm3} provided that $f^*$ is contained in the RKHS associated with the Gaussian kernel. This RKHS is fairly large as the Gaussian kernel is known to be universal in the sense that any continuous function can be well approximated by some function in the induced RKHS under the infinity norm \cite{Steinwart2008}. The above theoretical results can be further refined when $f^*$ belongs to some specific RKHS, and {some theoretical examples are provided in Section \ref{theoreticalexample:1}}.

	\section{{Theoretical examples}}\label{theoreticalexample:1}
	
	{This section provides some theoretical examples to illustrate the proposed method with the linear and quadratic  kernels. Moreover, we also discuss some possible treatments to improve the theoretical results with some additional technical assumptions.}
	
	\subsection{Linear kernel}\label{sec:4.1}
	
	Variable selection for linear model is of great interest in statistical literature due to its simplicity and interpretability. Particularly, the true regression function is assumed to be a linear function, $f^*(\bx)=\beta_0 + \bx^T\bfbeta^*$, and the true active set is defined as ${\cal A}^*=\{l: \beta^*_l \neq 0\}$. We also centralize the response and each variable, so that $\beta_0$ can be discarded from the linear model for simplicity.

	
	We now apply the general results in Section 3  to establish the sparsistency of the proposed algorithm under the linear model. We first scale the original data as {$\widetilde{\by}=p_n^{-1/2} \by$ and $\widetilde{\bx} = p_n^{-1/2} \bx$, and let ${\cal H}_K$ be the RKHS induced by the scaled linear kernel $K(\widetilde{\bx},\widetilde{\bu})=\widetilde{\bx}^T\widetilde{\bu}=p_n^{-1} \bx^T \bu$.  Then the true regression function can be rewritten as $f^*(\widetilde \bx)=\widetilde \bx^T  \bfbeta^*$.} With the scaled data, the ridge regression formula in \eqref{eqn:e1} becomes 
	\begin{align}
	\label{linear:1}
	{\widehat{\bfbeta} =\argmin_{{\bfbeta}} \ \frac{1}{n} \sum_{i=1}^n (\widetilde y_i - {\bfbeta}^T \widetilde{\bx}_i)^2 + p_n^{-1} \lambda_n\|{\bfbeta}\|^2.}
	\end{align}
	By the representer theorem, the solution of (\ref{linear:1}) is 
	\begin{align}
	\label{linear2}
	{
		\widehat{\bfbeta}={\bX}^T \big ( {\bX}{\bX}^T+ n \lambda_n {\bI}_n \big )^{-1} {\by},
	}
	\end{align}
	where ${\bX}=({\bx}_1,..., {\bx}_n)^T$ and ${\by}=( y_1,...,  y_n)^T$. It is equivalent to the standard formula for the ridge regression $\widehat{\bfbeta}=\big ( {\bX}^T{\bX}+ n  \lambda_n \bI_n \big )^{-1}{\bX}^T {\by}$ according to the Sherman-Morrison-Woodbury formula \cite{WangX2016}. {If we let $\lambda_n=0$, the estimate in (\ref{linear2}) is exactly the same as the HOLP estimate in \cite{WangX2016}. In other words, the HOLP method can be regarded as a special case of our proposed algorithm with the RKHS induced by the linear kernel}.
	
	The following Corollary \ref{Corollary:1} is a direct application of Theorem \ref{thm1} under the linear kernel.
	\begin{Corollary}\label{Corollary:1}
		Suppose that Assumptions S1 in the supplementary material is met. {Let ${\lambda}_n=O(p^{1/3}_nn^{-(1+\tau_1)/3} (\log n)^{2/3} )$,} then for any $\delta_n\geq 4(\sigma^2+\|{\bfbeta}^*\|_2^2)(\log n)^{-2}$, there exists some positive constant ${c_3}$ such that, with probability at least $1-\delta_n$, there holds
		\begin{align*}
		\| {\widehat{\bfbeta}} - {\bfbeta}^*  \| \leq c_3 \log \Big ( \frac{4}{\delta_n} \Big ) p_n^{1/6}n^{-\frac{1-2\tau_1}{6}} (\log n)^{{1}/{3}}.
		\end{align*}
		Additionally,	suppose that Assumption S2 in the supplementary material is met.  If 
		let {$v_n=\frac{s_1}{2}p_n^{1/6}n^{-\frac{1-2\tau_1}{6}} (\log n)^{\xi_2}$}, then we have
		\begin{align*}
		P\left( \widehat{\cal A}_{v_n}={\cal A}^* \right)\rightarrow 1, \ \ \mbox{as} \ \ n\rightarrow \infty,
		\end{align*}
		where $s_1$ and $\xi_2$ are provided in Assumption S2.
	\end{Corollary}
	
	Note that Corollary \ref{Corollary:1} holds when $p_n$ diverges at order ${o( {\min\{n^{{1-2\tau_1}}(\log n)^{-6\xi_2}, n^{{1+\tau_1}}(\log n)^{-2} \} } ).}$ Particularly, {when $\tau_1$ is sufficiently small, $p_n$ can diverge at the polynomial rate $o(n)$. This result is comparable with that in Shao and Deng \cite{Shao2012} under the finite second moment error assumption. The strong convergence rate obtained in Corollary \ref{Corollary:1} is also comparable with that in Theorem 2 of \cite{Shao2012}, and similar result holds for the required minimal signal strength.  }
	
	{\bf Remark 4:} Note that the proposed algorithm requires $f^* \in{\cal H}_K$, and thus $\|\bfbeta^*\|$ needs to be bounded, which implies that $p_0$ should be fixed in the linear case. Interestingly, if we take  $\lambda_n=0$ and all the technical assumptions stated in  \cite{WangX2016} are met, including that $\bx$ follows a spherically symmetric distribution and the noise $\epsilon$ has q-exponential tail, we can directly apply the theoretical results of the HOLP method to establish a similar selection consistency in Corollary \ref{Corollary:1}. As a direct consequence, $p_n$ and $p_0$ are allowed to diverge at some exponential and polynomial rate of $n$, respectively. 

	%
	
	
	\subsection{Quadratic kernel} 
	
	Variable selection for quadratic model is of great interest in statistical literature  \cite{Hao2014,  Kong2017, She2018}, where the true regression function is assumed to be  $f^*(\bx)=\beta_0+ \sum_{l=1}^{p_n} \beta^*_lx^l + \sum_{l \leq k} \gamma^*_{lk}x^lx^k$, where $\gamma^*_{lk}$'s are the true interaction coefficients and $\gamma^*_{lk}\neq 0$ implies that $x^l$ and $x^k$ have an interaction effect. The true active set is defined as
	\begin{align*}
	{\cal A}^*=\big \{l :  |\beta^*_l|+\sum_{k=1}^{p_n}|\gamma^*_{lk}| > 0  \big \},
	\end{align*}
	which contains variables contributing to $f^*$ through either the main factors or the interaction terms.	For simplicity, we denote $\overline{\bx}=(1, \sqrt{2}x_1,...,\sqrt{2}x_{p_n}, x_1^2,\sqrt{2}x_1x_2,...,\sqrt{2}x_1x_{p_n},$ $x_2^2,\sqrt{2}x_2x_3,....,x_{p_n}^2)^T$, and $\bftheta^*=(\beta^*_0,\bfbeta^{* T},\bfgamma^{* T})^T$ with $\bfbeta^*=(\beta^*_1,...,\beta^*_{p_n})^T/\sqrt{2}$ and $\bfgamma^*=(\gamma^*_{11},\gamma^*_{12}/\sqrt{2}...,\gamma^*_{22},\gamma_{23}^*/\sqrt{2},...,$ $\gamma_{(p_n-1)p_n}^*/\sqrt{2}, \gamma_{p_np_n}^*).$ Then, we scale the original data as $\check{\by}=p_n^{-1} \by$ and $\check{\bx} = p_n^{-1} \overline{\bx}$, and let ${\cal H}_K$ be the RKHS induced by a scaled quadratic kernel $K( \bx,  \bu)=(1+\bx^T\bu)^2/p^2_n= {\check{\bx}}^T \check{\bu}$. The true regression model can be rewritten as $f^*(\check \bx)=\check{\bx}^T  \bftheta^*$. Note that the quadratic model can be transformed into a linear form, then the established results in Section \ref{sec:4.1} can be directly applied. Specifically, with the scaled data, the ridge regression formula in \eqref{eqn:e1} becomes 
	\begin{align}\label{Quad:1}
	\widehat{\bftheta}=\argmin_{\bftheta} \ \frac{1}{n} \sum_{i=1}^n (\check{y}_i-\bftheta^T \check{\bx}_i)^2 +p_n^{-2}\lambda_n\|\bftheta\|^2.
	\end{align}
	Then the estimated active set is defined as  $
	\widehat{\cal A}_{v_n}=\big \{l :  | \widehat{\beta}_l|+ \sum_{k=1}^{p_n} |\widehat{\gamma}_{lk} | > v_n  \big \}
	$ with some pre-specified thresholding value $v_n$.
	
	With some slight modification of the proof of Corollary \ref{Corollary:1}, we obtain the following convergence results for the scaled quadratic kernel.
	\begin{Corollary}\label{Corollary:2}
		Suppose that Assumptions S3 in the supplementary material  is met. {Let {${\lambda}_n=O(p^{2/3}_nn^{-(1+\tau_2)/3} (\log n)^{2/3} )$},} then for any {$\delta_n\geq 4(\sigma^2+\|{\bftheta}^*\|_2^2)(\log n)^{-2}$}, there exists some positive constant ${c_4}$ such that, with probability at least $1-\delta_n$, there holds
		\begin{align*}
		\| {\widehat{\bftheta}} - {\bftheta}^*  \| \leq c_4 \log \Big ( \frac{4}{\delta_n} \Big ) p_n^{1/3}n^{-\frac{1-2\tau_2}{6}} (\log n)^{{1}/{3}}.
		\end{align*}
		Additionally,	suppose that Assumption S4 in the supplementary material is met. if let {$v_n=\frac{s_2}{2}p_n^{1/3}n^{-\frac{1-2\tau_2}{6}} (\log n)^{\xi_3}$}, then we have
		\begin{align*}
		P\left( \widehat{\cal A}_{v_n}={\cal A}^* \right)\rightarrow 1, \ \ \mbox{as} \ \ n\rightarrow \infty,
		\end{align*}
		where $s_2$ and $\xi_3$ are provided in Assumption S4.
	\end{Corollary}
	
	Note that the treatment in this subsection can be further extended to the polynomial regression model with degree $d$ by using the scaled polynomial kernel $K(\bx,\bu)=(1+\bx^T\bu)^d/p^d_n$, and similar theoretical results can be established for the proposed algorithm with the scaled polynomial kernel.

	\section{An extension: interaction selection}\label{Sec4}
	
	We now extend the proposed method to identify the truly informative interaction effects. In literature, a number of attempts have been made to identify the true interaction effects in both parametric and nonparametric regression models \cite{Lin2006, Choi2010, RadchenkoH2010, Hao2014, Hao2018}. Yet, most existing methods require some pre-specified working models and some of them are computationally demanding. {For example, the COSSO method \cite{Lin2006} and the SpIn method \cite{RadchenkoH2010} assume a second-order additive structure and need to enumerate $O(p^2)$ two-way interaction terms in the model, making their methods feasible only when $p$ is relatively small.} By contrast, our method can be extended directly and provide an efficient alternative for interaction selection without explicit model assumption. 
	
	Following the idea in Section 2, the true interaction effects can be defined as those with nonzero second-order gradient function $g^*_{lk}(\bx) = \partial^2 f^*(\bx) / \partial x^l \partial x^k$. Specifically, given the true active set ${\cal A}^*$, we denote
	\begin{align*}
	{\cal A}^*_{2}=\big \{l \in {\cal A}^*:  \|g^*_{lk}\|_2 > 0,~\mbox{for some}\  k \in {\cal A}^* \big \},
	\end{align*}
	which contains the variables that contribute to the interaction effects in $f^*$. Further, let ${\cal A}_1^*={\cal A}^* \setminus {\cal A}^*_2$, which contains the variables that contribute to the main effects of $f^*$ only. 
	
	Therefore, the main goal of interaction selection is to correctly estimate both ${\cal A}^*_1$ and ${\cal A}^*_2$. First, let $K(\cdot,\cdot)$ be a forth-order differentiable kernel function, then it follows from Lemma 1 in   the supplementary material that for any $f \in {\cal H}_K$,
	$$
	g_{lk}(\bx)=\frac{\partial^2 f(\bx)}{\partial x^l \partial x^k} =\langle f, {\partial_{lk} {K}_{\bx}} \rangle_K \leq \|\partial_{lk} K_{\bx}\|_K\| f\|_K,
	$$ 
	where $ {\partial_{lk} {K}_{\bx}}=\frac{\partial^2 K(\bx,\cdot)}{\partial x^l \partial x^k}$. Then, given $\widehat f$ from \eqref{eqn:e1}, its second-order gradient function is
	$$
	\widehat{g}_{lk}(\bx) =\frac{\partial^2 \widehat{f}(\bx)}{\partial x^l \partial x^k}=   \widehat{\bfalpha}^T{\partial_{lk} {\bK}_n({\bx})},
	$$
	where $\partial_{lk}\bK_n(\bx)=\frac{\partial {\bK}_n({\bx})}{\partial x^l \partial x^k}$. Its empirical norm is $ \| \widehat{g}_{lk}  \|^2_n = \frac{1}{n}\sum_{i=1}^n  \big ( \widehat{g}_{lk}({\bx}_i) \big )^2$. With some pre-defined thresholding value $v_n^{int}$, the estimated ${\cal A}^*_1$ and ${\cal A}^*_2$ are set as
	$$
	\widehat{\cal A}_{2}=\big \{l \in\widehat{\cal A}:  \| \widehat{g}_{lk}  \|^2_n >v_n^{int},~\mbox{for some}\ k \in \widehat{\cal A} \big \}  \ \mbox{and} \ \widehat{\cal A}_1= \widehat{\cal A}\setminus \widehat{\cal A}_{2},
	$$
	respectively. The following technical assumption is made to establish the interaction selection consistency for the proposed method.
	
	\noindent{\bf Assumption 5}: There exists some constant $\kappa_3$ such that $\sup_{\bx \in {\cal X}} \|\partial_{lk} K_{\bx}\|_{K}\leq \kappa_3$ for any $l$ and $k$. 
	
	Assumption 5 can be regarded as the extension of Assumptions 2 by requiring 
	the boundedness of the second-order gradients of $K_{\bx}$.

	\begin{theorem}\label{inter:thm4}
		Suppose the assumptions of Theorem \ref{thm3} and Assumption 5 are met. Let $P(\widehat{\cal A}\neq{\cal A}^*)=\Delta_n$. Then with probability at least $1-\delta_n-\Delta_n$, there holds
		$$
		\max_{l,k \in \widehat{\cal A}} \ \big|  \| \widehat{g}_{lk}   \|^2_n - \| g^*_{lk}\|_2^2  \big | \leq  b_{n,2} \max \{ \kappa_1\|f^*\|_K, q^{-1}(\log\frac{4c_1 n}{\delta_n}) \}  \log\Big ( \frac{8p_0^2}{\delta_n} \Big ) n^{-\frac{2r-1}{2(2r+1)}}  ,
		$$
		where $b_{n,2}=4\max\{\kappa^2_3, \|f^*\|^2_{K}, \kappa^2_3\|f^*\|_{K}\} \max\{3\kappa_1,  2\sqrt{2}\kappa_3^2, \|L_K^{-r}f^*\|_2\}$.
	\end{theorem}
	Theorem \ref{inter:thm4} shows that $\left \|\widehat{g}_{lk}\right \|^2_n$ converges to $\left \| g_{lk}^* \right \|_{2}^2$ with high probability, which is crucial to establish the interaction selection consistency.  
	
	\noindent{\bf Assumption 6}: There exists some positive constant $\xi_4< \frac{2r-1}{2(2r+1)}$ such that 
	$\min_{\substack{l,k \in {\cal A}^*_{2}} } \|g^*_{lk}\|_2^2 > b_{n,2} \max \{ \kappa_1\|f^*\|_K, q^{-1}(\log\frac{4c_1 n}{\delta_n})    \}  n^{-\xi_4}  \log p_0$.
	
	Assumption 6 can be regarded as the extension of Assumption 3 by requiring  the true second-order gradient functions have sufficient information about the interaction effects.  
	 
	\begin{theorem}\label{inter:thm5}
		Suppose the assumptions of Theorem \ref{inter:thm4} as well as Assumption 6 are met. By taking $v^{int}_n=\frac{b_{n,2}}{2} \max \{ \kappa_1\|f^*\|_K, q^{-1}(\log\frac{4c_1 n}{\delta_n}) \} n^{-\xi_4}  \log p_0$, we have
		\begin{align*}
		P \Big ( \widehat{\cal A}_{2}={\cal A}_{2}^*, \widehat{\cal A}_1={\cal A}_1^* \Big )\rightarrow 1, \ \ \mbox{as} \ \ n\rightarrow \infty.
		\end{align*}
	\end{theorem}
	Theorem \ref{inter:thm5} shows that the proposed interaction selection method can exactly detect the interaction structure with probability tending to 1. Note that this result is established without requiring the strong heredity assumption, which is often assumed by the existing parametric interaction selection methods \cite{Choi2010, Hao2014}. It is also clear that the proposed method can be extended to detect higher-order interaction effects, which is of particular interest in some real applications \cite{Ritchie2001}.  

	\section{Numerical experiments}\label{Sec5}
	
	In this section, the numerical performance of the proposed method is examined, and compared against some existing methods, including the distance correlation learning \cite{Szekely2007} and the quantile-adaptive screening \cite{Hexm2013}. As these two methods are designed for screening only, they are also truncated by some thresholding values to conduct sparse learning. For simplicity, we denote these three methods as GM, DC-t and QaSIS-t, respectively. Note that the computational cost of most existing gradient-based methods \cite{Rosasco2013, YangL2016} can be very expensive, and thus they are not included in the numerical comparison with large dimension. 
	
	{In all the simulation examples, no prior knowledge about the true regression function is assumed, and the Gaussian kernel $K(\bu,\bv) = \exp \big (-\|\bu-\bv\|^2 / 2\sigma_n^2 \big )$ is used to induce the RKHS, where $\sigma_n$ is set as the median of all the pairwise distances among the training sample. For the proposed method, we set the ridge parameter $\lambda_n=0.001$ in all simulated examples, and use the stability criterion in Section \ref{Tune} to conduct a grid search for the optimal thresholding parameter $v_n$, where the grid is set as $\{10^{-3+0.1s} : s = 0, ..., 60\}$}. 
	
	\subsection{Simulated examples}
	
	Two simulated examples are examined under various scenarios. 
	
	
	{\noindent\bf Example 1:} We first generate $x_i = (x_{i1}, ..., x_{i{p}})^T$ with $x_{ij} = \frac{W_{ij} +\eta U_i}{1+\eta}$, where $W_{ij}$ and $U_i$ are independently drawn from $U(-0.5, 0.5)$. The response $y_i$ is generated as $y_i=f(\bx_i)+\epsilon_i,$ where $f^*(\bx_i)=6f_1(x_{i1})+4f_2(x_{i2})f_3(x_{i3})+6f_4(x_{i4})+5f_5(x_{i5})$,
	with $f_1(u)=u, f_2(u)=2u+1, f_3(u)=2u-1, f_4(u)=0.1\sin(\pi u)+0.2\cos(\pi u)+0.3(\sin(\pi u))^2+0.4(\cos(\pi u))^3+0.5(\sin(\pi u))^3$, $f_5(u)=\sin(\pi u)/(2-\sin(\pi u))$, and $\epsilon_i$'s are independently drawn from $N(0,1)$. Clearly, the first $5$ variables are truly informative.
	
	{\noindent\bf Example 2:} The generating scheme is similar to Example 1, except that $W_{ij}$ and $U_i$ are independently drawn from $U(0, 1)$ and $f^*(\bx)=20x_1x_2x_3+5x_4^2+5x_5$. The first $5$ variables are truly informative.
	
	For each example, we  consider scenarios with $(n, p) = (400, 500), (400, 1000), (500, 10000),$ $(500, 50000)$ and $(500,100000)$. For each scenario, $ \eta= 0$ and $ 1$ are examined. When $\eta = 0$, the variables are completely independent, whereas when $\eta=1$, correlation structure are added among the variables. Each scenario is replicated $50$ times. The averaged signal-to-noise ratios (SNR) of  the simulated examples under different scenarios are summarized in Table \ref{tab:sim0}. The averaged performance measures are summarized in Tables \ref{tab:sim1} and \ref{tab:sim2}, where Size is the averaged number of selected informative variables, TP is the number of truly informative variables selected, FP is the number of truly uninformative variables selected, C, U, O are the times of correct-fitting, under-fitting, and over-fitting, respectively. 
	\begin{center} 
			\begin{tabular}[t]{c}
			\hline
			\hline
			Tables \ref{tab:sim0} -- \ref{tab:sim2} about here\\
			\hline
			\hline
		\end{tabular}
	\end{center}
Clearly, the SNRs of the simulated examples are comparable to those
in \cite{Lin2006, Huang2010}. It is evident that GM outperforms the other methods in both examples. In Example 1, GM is able to identify all the truly informative variables in most replications. However, the other two methods tend to miss some truly informative variables, probably due to the interaction effect between $x^2$ and $x^3$. In Example 2, with a three-way interaction term involved in $f^*(\bx)$, GM is still able to identify all the truly informative variables with high accuracy, but the other two methods tend to underfit by missing some truly informative variables in the interaction term.  {It is also interesting to notice that GM tends to overselect variables in some cases, which is generally less severe than under-selecting truly informative variables. 
		
	Note that if we do not threshold DC and QaSIS, they tend to overfit almost in every replication as both screening methods tend to keep a substantial amount of uninformative variables to attain the sure screening property. Furthermore, when the correlation structure with $\eta=1$ is considered, identifying the truly informative variables becomes more difficult, and both DC-t and QaSIS-t become unstable, yet GM still outperforms these two competitors and can exactly identify all the truly informative variables in most replications.
		
		
		\subsection{Supermarket dataset}
		
		We now apply the proposed method to a supermarket dataset in Wang \cite{Wang2009}. The dataset is collected from a major supermarket located in northern China, consisting of daily sale records of $p=6398$ products on $n=464$ days. In this dataset, the response is the number of customers on each day, and the variables are the daily sale volumes of each product. The primary interest is to identify the products whose sale volumes are related with the number of customers, and then to design sale strategies based on those products. The dataset is pre-processed so that both the response and predictors have zero mean and unit variance.
		
		In addition to GM, DC-t and QaSIS-t, we also include the comparisons with  SCAD \cite{Fan2001} and MCP \cite{ZhangC2010}. As the truly informative variables are unknown for the supermarket dataset, we report the prediction performance of each method. Specifically, the supermarket dataset is randomly split into two parts, with 164 observations for testing and the remaining for training. We first apply each method to the full dataset to select the informative variables, and then 	 refit a kernel ridge regression model  for the nonparametric methods and a linear ridge regression for the parametric methods with the selected variables on the training set. The prediction performance of each ridge regression model is measured on the testing set. The procedure is replicated 1000 times, and the number of selected variables, the averaged prediction errors and the out of sample $R^2$ the are summarized in Table \ref{tab:sim3}.
		\begin{center}
			\begin{tabular}[t]{c}
				\hline
				\hline
				Table \ref{tab:sim3} is about here \\
				\hline
				\hline
			\end{tabular}
		\end{center}

		As Table \ref{tab:sim3} shows, GM selects 10 variables, whereas DC-t and QaSIS-t select 7 variables and SCAD and MCP select 59 and 28 variables, respectively. The average prediction error of GM is smaller than that of the other four methods, implying that DC-t and QaSIS-t may miss some truly informative variables that deteriorate their prediction accuracy, and SCAD and MCP may include too many noise variables.  Precisely, among the 10 selected variables by GM, $X^{14}, X^{18}, X^{42}, X^{56}$ and $X^{75}$ are missed by both DC-t and QaSIS-t. The scatter plots of the response against these five variables are presented in Figure \ref{fig:20}.

		\begin{center}
			\begin{tabular}[t]{c}
				\hline
				\hline
				Figure \ref{fig:20} is about here \\
				\hline
				\hline
			\end{tabular}
		\end{center}
		It is evident that the response and these variables have shown some clear relationship, which supports the advantage of GM in identifying the truly informative variables.
		
	\section{Summary}\label{Sec6}
    This article proposes a novel gradient-based sparse learning method, which can simultaneously enjoy methodological flexibility, numerical efficiency and asymptotic consistency. It provides a novel and promising way to conduct sparse learning for nonparametric models. The proposed method is simple and efficient in that the kernel ridge regression has analytic solution and the estimated gradient functions can be directly computed by using the derivative reproducing property \cite{ZhouD2007}. It can be easily scaled up to analyze datasets with huge dimensions. The theoretical results are established without requiring any restrictive model assumption, which justifies the robustness of the proposed method to the underlying data distribution.

    One interesting future work is to consider a more general scenario with $f^*$ out of the specified RKHS ${\cal H}_K$, such as a non-differentiable $f^*$. One possible remedial route is to consider the true active set ${\cal A}^*=\{l: D_l(f^*)>0\}$, where ${D}_{l}(f^*)=  \max_{{\bx}^{-l}}\big | \max_{x^l}f^*(x^l,{\bx}^{-l})-\min_{x^l}f^*(x^l,{\bx}^{-l}) \big | >0$ measures the largest possible change of $f(\bx)$ along $x^l$, and ${\bx}^{-l}$ denotes all variables except for $x^l$. Then the equivalence between ${D}_{l}(f^*)$ and the gradients of some intermediate function $f^0 \in {\cal H}_K$ is  to be examined in order to bridge the gap between $f^*$ and ${\cal H}_K$. Another interesting  future work is to  extend the proposed method to deal with  mixed-type  predictors, and ${D}_{l}(f^*)$ can also be used to measure the significance of each variable.

	    \section*{Supplementary materials}
		
		Proofs of Theorems 3 and 4, some necessary lemmas and their proofs as well as  verification of the theoretical examples are provided in the supplementary materials.
 
		\section*{Acknowledgment}
        Xin He’s research was supported in part by NSFC-11901375 and Shanghai Pujiang Program 2019PJC051.  Junhui Wang’s research was supported in part by HK RGC Grants GRF-11303918  and GRF-11300919.  Shaogao Lv's research was partially supported by NSFC-11871277. The authors also thank the associate editor and two anonymous  referees for their constructive suggestions.
		
		\section*{Appendix: technical proof} 
		
				
		{\noindent\bf Proof of Theorem \ref{thm1}.} For simplicity, we denote two events
     $$
     {\cal C}_1 = \left \{ {\cal Z}^n: \  \|\widehat{f}-{f}^*  \|_{K} \geq 2\log{\frac{8}{\delta_n}} \left ( \frac{3\kappa_1 }{{n}^{1/2}\lambda_n} \big (\kappa_1\|f^*\|_K+ q^{-1}(\log\frac{2c_1 n}{\delta_n}) \big ) + \lambda_n^{r-{1}/{2}} \|L^{-r}_{K}f^*\|_2 \right) \right \},
     $$
   $$
   {\cal C}_2 = \left \{ {\cal Z}^n: \  \max_{i=1,...,n}|y_i|>  \kappa_1\|f^*\|_K+ q^{-1}(\log\frac{2c_1 n}{\delta_n}) \right \},
   $$   
   and ${\cal C}_2^c$ denotes the complement of ${\cal C}_2$. Then $P({\cal C}_1)$ can be decomposed as
     	\begin{align*}
     	P({\cal C}_1)&= P\left ({\cal C}_1\cap {\cal C}_2 \right ) + P\left ({\cal C}_1\cap  {\cal C}^c_2 \right )\leq  P\left ( {\cal C}_2\right ) + P\left ({\cal C}_1 \mid {\cal C}_2^c\right )=P_1+P_2.
     	\end{align*} 
     	For $P_1$, 
by Assumption 3, we have 
     	\begin{align}\label{eqn:tail}
     	P(\max_{i=1,...,n}|\epsilon_i| \geq t) & = P (\cup_{i=1}^n|\epsilon_i| \geq t) \leq n  P (|\epsilon_i| \geq t) \leq c_1 n\exp\{-q(t)\}.
     	\end{align}
     	By Assumption 1 and \eqref{eqn:tail}, for any $\delta_n \in (0,1)$, with probability at least $1-\frac{\delta_n}{4}$, there holds
     	$$
     	\max_{i=1,...,n}|y_i|\leq \kappa_1\|f^*\|_K+\max_{i=1,...,n}|\epsilon_i|\leq  \kappa_1\|f^*\|_K+q^{-1}(\log\frac{4c_1 n}{\delta_n}),
     	$$
     	implying that $P\left ( {\cal C}_2\right )\leq \frac{\delta_n}{4}$.
     		
		For $P_2$, note that 
		$$
		\|\widehat{f} - f^*\|_K \leq \|\widehat{f} - \widetilde{f} \|_K+ \|\widetilde{f} - f^*\|_K.
		$$
		We first bound $\|\widetilde{f}-f^*\|_{K}$ following the similar treatment as in Smale and Zhou \cite{Smale2005}. Suppose $\{\mu_i,e_i\}_{i\geq 1}$ are the normalized eigenpairs of the integral operator $L_K: {\cal L}^2( {\cal X} ,{\rho_{\bx}})\rightarrow {\cal L}^2( {\cal X} ,{\rho_{\bx}})$, we have 
		$$
		L^{1/2}_Ke_i=\sum_{j\geq 1}\mu_j^{1/2}\langle e_i, e_j\rangle_2e_j=\mu^{1/2}_ie_i\in {\cal H}_K,
		$$ and 
		$$
		\|\mu^{1/2}_i e_i \|_{K}=\Big (  \sum_{j\geq 1} \frac{\langle \mu_i^{1/2}e_i, e_j \rangle^2_2}{\mu_j}\Big )^{1/2}= \langle e_i, e_i \rangle_2=1,
		$$ 
		when $\mu_i>0$. Thus by Assumption 1, there exists some function $h=\sum_{i\geq 1}\langle h , e_i \rangle_2e_i \in {\cal L}^2({\cal X},\rho_{\bx})$ such that $f^*=L^r_K h =\sum_{i\geq 1}\mu^r_i\langle h , e_i \rangle_2e_i \in {\cal H}_K$.
		Directly calculation yields to
		\begin{align*}
		\widetilde{f} - f^*&=\big ( L_K + \lambda_n I \big )^{-1}  L_{K} {f}^*  - f^* = \big ( L_K + \lambda_n I \big )^{-1} \big ( -\lambda_n f^* \big ) \\
		&=  - \sum_{i\geq 1}\frac{\lambda_n}{\lambda_n+\mu_i}\mu^{r}_i \langle h , e_i \rangle_2e_i.
		\end{align*}
		Therefore, the RKHS-norm of $\widetilde{f}- f^*$ can be bounded as
		\begin{align}\label{eqn:zhou4}
		\big \| \widetilde{f}- f^* \big \|^2_{K} &=    \sum\nolimits_{i\geq 1} \big ( \frac{\lambda_n}{\lambda_n+\mu_i}\mu_i^{r-{1}/{2}}\langle h , e_i \rangle_2  \big )^2  \|\mu^{{1}/{2}}_ie_i\|^2_{K} \nonumber\\
		&= \sum\nolimits_{i\geq 1} \big ( \frac{\lambda_n}{\lambda_n+\mu_i}\mu_i^{r-{1}/{2}}\langle h , e_i \rangle_2  \big )^2 \nonumber \\
		& = \lambda_n^{2r-1} \sum\nolimits_{i\geq 1} \Big ( \frac{\lambda_n}{\lambda_n+\mu_i} \Big)^{3-2r} \Big ( \frac{\mu_i}{\lambda_n+\mu_i} \Big )^{2r-1} \langle h , e_i \rangle^2_2 \nonumber \\
		&\leq \lambda_n^{2r-1}\sum\nolimits_{i\geq 1}  \langle h , e_i \rangle^2_2 = \lambda_n^{2r-1} \|h\|_2^2 = \lambda_n^{2r-1} \|L_K^{-r}f^*\|^2_2. 
		\end{align}
		It then follows from Proposition 1 in the supplemental material that
		$$
		P_2 \leq P \Big (\| \widehat{f}-\widetilde{f}  \|_{K} \geq \log \frac{8}{\delta_n} \frac{6\kappa_1 }{\lambda_n n^{1/2} }   (\kappa_1\|f^*\|_K+q^{-1}(\log\frac{4c_1 n}{\delta_n})) \mid {\cal C}_2^c \Big ) \leq {\delta_n}/{4}.
		$$
		
		Combining the upper bounds of $P_1$ and $P_2$ yields that $P({\cal C}_1)\leq {\delta_n}/{4} +{\delta_n}/{4} \leq \delta_n/2$.  
		Thus, with probability at least $1-\delta_n/2$, there holds
		$$
		\|\widehat{f}-{f}^*  \|_{K} \leq 2\log{\frac{8}{\delta_n}} \left ( \frac{3\kappa_1  }{{n}^{1/2}\lambda_n} (\kappa_1\|f^*\|_K+q^{-1}(\log\frac{4c_1 n}{\delta_n})) + \lambda_n^{r-{1}/{2}} \|L^{-r}_{K}f^*\|_2 \right).
		$$
		
		Now we turn to establish the weak convergence rate of $\widehat{g}_l$ in estimating $g_l^*$. We first introduce some notations. Define the sample operators for gradients $\widehat{D}_l:{\cal H}_K\rightarrow {\cal R}^n$ and their adjoint operators $\widehat{D}_l^*:{\cal R}^n\rightarrow {\cal H}_K$ as
		$$
		( \widehat{D}_lf )_i=\big \langle f, \partial_l K_{{\bx}_i}\rangle_K\  ~\mbox{and}~\  \widehat{D}_l^*\bc= \frac{1}{n}\sum_{i=1}^n \partial_l K_{{\bx}_i}c_i,
		$$
		respectively. And the integral operators for gradients $D_l: {\cal H}_K\rightarrow {\cal L}^2(\rho_{\bx},{\cal X})$ and $D^*_l: {\cal L}^2(\rho_{\bx},{\cal X})\rightarrow {\cal H}_K$ are defined as
		$$
		D_lf=\langle f, \partial_l K_{\bx} \rangle_K\  ~\mbox{and}~\  D_l^*{f}=\int \partial_lK_{\bx}f(\bx)d\rho_{\bx}.
		$$
		Note that $D_l$ and $\widehat{D}_l$ are the Hilbert-Schimdt operators by Propositions $12$ and $13$ of Rosasco et al. \cite{Rosasco2013}, then we have
		$$
		D^*_lD_{l}f=\int \partial_l{K}_{\bx} g_l(\bx) d \rho_{\bx}\ ~\mbox{and}~\  \widehat{D}^*_l\widehat{D}_{l}f=\frac{1}{n}\sum_{i=1}^n \partial_l{K}_{{\bx}_i} g_l({\bx}_i).
		$$
		Furthermore, we denote $HS(K)$ as a Hilbert space with all the Hilbert-Schmidt operators on ${\cal H}_K$, which endows with a norm 
		$\|\cdot\|_{HS}$ such that $\|T\|_K\leq \|T\|_{HS}$ for any $T\in HS(K)$. 
		
		
		With these operators, simple algebra yields that
		\begin{align} 
		& \big | \| \widehat{g}_{l} \|^2_n -  \| g^*_l \|_{2}^2 \big | \nonumber\\
		&=  \Big | \frac{1}{n}\sum\limits_{i=1}^n\left( \widehat{g}_l({\bx}_i) \right)^2 -  \int \left ( g^*_l(\bx)\right)^2 d\rho_{\bx} \Big | \nonumber  \\
		& =   \Big |  \frac{1}{n}\sum\limits_{i=1}^n \widehat{g}_l({\bx}_i) \big \langle \widehat{f}, \partial_l{K}_{{\bx}_i} \big \rangle_K   -  \int  g^*_l(\bx)\left \langle f^*, \partial_l{K}_{\bx} \right  \rangle_K d\rho_{\bx} \Big | \nonumber \\
		& =  \Big | \big \langle \widehat{f}, \frac{1}{n}\sum\limits_{i=1}^n\widehat{g}_l({\bx}_i) \partial_l{K}_{{\bx}_i} \big  \rangle_K   - \big \langle f^*,  \int  g^*_l(\bx)\partial_l{K}_{\bx} d\rho_{\bx} \big \rangle_K \Big |  \nonumber \\
		& =  \Big |  \big \langle \widehat{f} - f^*,  \widehat{D}^{*}_l\widehat{D}_l\widehat{f} \big  \rangle_K   + \big \langle f^*,  \widehat{D}^{*}_l\widehat{D}_l(\widehat{f}-f^*) \big \rangle_K + \big \langle f^*,  (\widehat{D}^{*}_l\widehat{D}_l - {D}^{*}_l{D}_l )f^* \big \rangle_K \Big | \nonumber\\
		&=   \Big |  \big \langle \widehat{f} - f^*,  \widehat{D}^{*}_l\widehat{D}_l(\widehat{f}-f^*) \big  \rangle_K+  \big \langle\widehat{D}^{*}_l\widehat{D}_l f^*,  \widehat{f}-f^* \big \rangle_K   + \nonumber \\
		& \hspace{4.2cm}\big \langle f^*,  \widehat{D}^{*}_l\widehat{D}_l(\widehat{f}-f^*) \big \rangle_K + \big \langle f^*,  (\widehat{D}^{*}_l\widehat{D}_l -   {D}^{*}_l{D}_l )f^* \big \rangle_K  \Big | \nonumber \\
		&   \leq \|\widehat{f} - f^*\|^2_K \|\widehat{D}^{*}_l\widehat{D}_l\|_{HS} + 2\|\widehat{f} - f^*\|_K\|f^*\|_K\|\widehat{D}^{*}_l\widehat{D}_l\|_{HS} + \nonumber \\
		& \hspace{8cm}\|\widehat{D}^{*}_l\widehat{D}_l -  {D}^{*}_l{D}_l\|_{HS} \|f^*\|_K^2 \nonumber,
		\end{align}
		where the last inequality follows from the Cauthy-Schwartz inequality. It then suffices to bound the terms in the upper bound of $\big | \| \widehat{g}_{l} \|^2_n -  \| g^*_l \|_{2}^2 \big |$ separately. Note that $\|f^*\|_K$ is a bounded quantity, and it follows from Assumption 2 and Rosasco et al. \cite{Rosasco2013} that $\max_l \big \| \widehat{D}^*_l\widehat{D}_{l} \big \|_{HS}=\max_l\|\partial_l{K}_{\bx} \|_K^2 \leq  \kappa_2^2$. Hence, we have 
		\begin{align*}
		&\max_{1 \leq l \leq p} \ \big | \| \widehat{g}_{l}  \|^2_n -   \| g^*_l  \|_{2}^2 \big | \\
		& \leq a_1\Big ( 
		\|\widehat{f}-f^*\|_{K}^2 + 2\|\widehat{f}-f^*\|_{K}  + \max_{1 \leq l \leq p} \ \|\widehat{D}^{*}_l\widehat{D}_l - {D}^{*}_l{D}_l \|_{HS}
		\Big ),
		\end{align*}
		where $a_1=\max\{\kappa^2_2, \kappa^2_2\|f^*\|_{K}, \|f^*\|^2_{K}\}$. When $ \| \widehat{f} - f^*\|_{K}$ is sufficiently small, the upper bound can be simplified to
		$$
		\max_{1 \leq l \leq p} \ \big | \| \widehat{g}_{l}  \|^2_n -   \| g^*_l  \|_{2}^2 \big | \leq a_1\Big (   3
		\|\widehat{f}-f^*\|_{K}  + \max_{1 \leq l \leq p}  \|\widehat{D}^{*}_l\widehat{D}_l - {D}^{*}_l{D}_l \|_{HS}
		\Big ),
		$$
		where $\|\widehat{f}-f^*\|_{K}$ is bounded in the first half of the proof. Furthermore, for any $\epsilon_n \in(0,1)$, by the concentration inequalities for $HS(K)$ \cite{Rosasco2013}, we have 
		\begin{align*}
		\ P \Big ( \big \| \widehat{D}^*_l\widehat{D}_{l}   -  D^*_lD_{l} \big \|_{HS} \geq \epsilon_n  \Big ) \leq 2p\exp \Big ( -\frac{n\epsilon_n^2}{8\kappa_2^4} \Big ),
		\end{align*}
		for any $l=1,\ldots,p$. Therefore, with probability at least $1-\delta_n/2$, there holds 
		$$
		\max_{1 \leq l \leq p} \ \big \| \widehat{D}^*_l\widehat{D}_{l}   -  D^*_lD_{l} \big \|_{HS} \leq \Big ( \frac{8\kappa_2^4}{n}\log\frac{4p}{\delta_n} \Big )^{{1}/{2}}.
		$$
		
		Combining all the upper bounds above, we have with probability at least $1-\delta_n$, there holds
		\begin{align*}
		&\max_{1 \leq l \leq p} \ \left| \left \| \widehat{g}_l  \right \|^2_n -  \left \| g^*_l \right \|_{2}^2   \right| \\ 
		&\leq 2 a_1 \Big ( 3\log{\frac{8}{\delta_n}} \big (  \frac{3\kappa_1}{{n}^{1/2}\lambda_n} (\kappa_1\|f^*\|_K+q^{-1}(\log\frac{4c_1 n}{\delta_n})) + \lambda_n^{r-{1}/{2}} \|L^{-r}_{K}f^*\|_2 \big ) + \big ( \frac{2\kappa_2^4}{n}\log\frac{4p}{\delta_n} \big )^{{1}/{2}} \Big ).
		\end{align*}
		This implies the desired results immediately with $\lambda_n=n^{-\frac{1}{2r+1}}$.  \hfill$\blacksquare$

		{\noindent\bf Proof of Theorem \ref{thm3}.} We first show that  ${\cal A}^* \subset \widehat{\cal A}$ in probability. If not, suppose there exists some $l' \in {\cal A}^*$ but $l' \notin \widehat{\cal A}$, and thus $  \| \widehat{g}_{l^{'}} \|^2_n\leq v_n$. By Assumption 4, we have with probability $1-\delta_n$ that
		\begin{align*}
		\big |  \| \widehat{g}_{l^{'}}   \|^2_n -   \| g^*_{l^{'}}  \|_{2}^2   \big |   \geq  \| g^*_{l^{'}}  \|_{2}^2   -     \| \widehat{g}_{l^{'}}  \|^2_n & > b_{n,1} \max \{ \kappa_1\|f^*\|_K, q^{-1}(\log\frac{4c_1 n}{\delta_n}) \}  n^{-\xi_1} \log p -v_n \\ &=\frac{b_{n,1}}{2} \max \{ \kappa_1\|f^*\|_K, q^{-1}(\log\frac{4c_1 n}{\delta_n}) \}  n^{-\xi_1} \log p, 
		\end{align*}
		which contradicts with Theorem \ref{thm1}. This implies that ${\cal A}^*\subset\widehat{\cal A}$ with probability at least $1-\delta_n$.
		
		Next, we show that $\widehat{\cal A} \subset {\cal A}^*$ in probability. If not, suppose there exists some $l' \in \widehat{\cal A}$ but $l' \notin {\cal A}^*$, which implies
		$  \|\widehat{g}_{l^{'}} \|^2_n> v_n$ but $  \| g^*_{l^{'}}  \|_{2}^2 = 0$, and then with probability at least $1-\delta_n$, there holds
		$$
		\big|  \| \widehat{g}_{l^{'}}  \|^2_n -   \| g^*_{l^{'}}  \|_{2}^2   \big |   > v_n= \frac{b_{n,1}}{2} \max \{ \kappa_1\|f^*\|_K, q^{-1}(\log\frac{4c_1 n}{\delta_n}) \}  n^{-\xi_1} \log p.
		$$
		This contradicts with Theorem \ref{thm1} again, and thus $\widehat{\cal A}\subset {\cal A}^*$ with probability at least $1-\delta_n$.
		Combining these two results yields the desired sparsistency.
		\hfill$\blacksquare$


		\bibliography{bibtex}

\begin{thebibliography}{10}

\bibitem{Barber2015}
R.~Barber and E.~Cand$\ddot{\text{e}}$s.
\newblock Controlling the false discovery rate via knockoffs.
\newblock {\em Annals of Statistics}, \textbf{43}:2055--2085, 2015.

\bibitem{Bartlett2002}
P.~Bartlett and S.~Mendelson.
\newblock Rademacher and gaussian complexities: risk bounds and structural
  results.
\newblock {\em Journal of Machine Learning Research}, \textbf{3}:463--482,
  2002.

\bibitem{Bondell2009}
H.~Bondell and L.~Li.
\newblock Shrinkage inverse regression estimation for model free variable
  selection.
\newblock {\em Journal of the Royal Statistical Society, Series B},
  \textbf{71}:287--299, 2009.

\bibitem{Choi2010}
N.~Choi, W.~Li, and J.~Zhu.
\newblock Variable selection with the strong heredity constraint and its oracle
  property.
\newblock {\em Journal of the American Statistical Association},
  \textbf{105}:354--364, 2010.

\bibitem{Dasgupta2018}
S.~Dasgupta, Y.~Goldberg, and M.~Kosorok.
\newblock Feature elimination in kernel machines in moderately high dimensions.
\newblock {\em Annals of Statistics}, \textbf{47}:497--526, 2019.

\bibitem{Fan2011}
J.~Fan, Y.~Feng, and R.~Song.
\newblock Nonparametric independence screening in sparse ultrahigh dimensional
  additive models.
\newblock {\em Journal of the American Statistical Association},
  \textbf{106}:544--557, 2011.

\bibitem{Fan2001}
J.~Fan and R.~Li.
\newblock Variable selection via nonconcave penalized likelihood and its oracle
  properties.
\newblock {\em Journal of the American Statistical Association},
  \textbf{96}:1348--1360, 2001.

\bibitem{FAN2008}
J.~Fan and J.~Lv.
\newblock Sure independence screening for ultrahigh dimensional feature space
  (with discussion).
\newblock {\em Journal of the Royal Statistical Society, Series B},
  \textbf{70}:849--911, 2008.

\bibitem{Fischer2017}
S.~Fischer and I.~Steinwart.
\newblock Sobolev norm learning rates for regularized least square algorithm.
\newblock {\em Manuscript}, 2019.

\bibitem{Fukumiza2014}
K.~Fukumizu and C.~Leng.
\newblock Gradient-based kernel dimension reduction for regression.
\newblock {\em Journal of the American Statistical Association},
  \textbf{109}:359--370, 2014.

\bibitem{Hao2018}
N.~Hao, Y.~Feng, and H.~Zhang.
\newblock Model selection for high dimensional quadratic regression via
  regularization.
\newblock {\em Journal of the American Statistical Association},
  \textbf{113}:615--625, 2018.

\bibitem{Hao2014}
N.~Hao and H.~Zhang.
\newblock Interaction screening for ultra-high dimensional data.
\newblock {\em Journal of the American Statistical Association},
  \textbf{109}:1285--1301, 2014.

\bibitem{Hexm2013}
X.~He, L.~Wang, and H.~Hong.
\newblock Quantile-adaptive model-free variable screening for high-dimensional
  heterogeneous data.
\newblock {\em Annals of Statistics}, \textbf{41}:342--369, 2013.

\bibitem{Huang2010}
J.~Huang, J.~Horowitz, and F.~Wei.
\newblock Variable selection in nonparametric additive models.
\newblock {\em Annals of Statistics}, \textbf{38}:2282--2313, 2010.

\bibitem{Kong2017}
Y.~Kong, D.~Li, Y.~Fan, and J.~Lv.
\newblock Interaction pursuit in high-dimensional multi-response regression via
  distance correlation.
\newblock {\em Annals of Statistics}, \textbf{45}:897--922, 2017.

\bibitem{LiB2005}
B.~Li, H.~Zha, and F.~Chiaromonte.
\newblock Contour regression: a general approach to dimension reduction.
\newblock {\em Annals of Statistics}, \textbf{33}:1580--1616, 2005.

\bibitem{Lin2006}
Y.~Lin and H.~Zhang.
\newblock Component selection and smoothing in multivariate nonparametric
  regression.
\newblock {\em Annal of Statistics}, \textbf{34}:2272--2297, 2006.

\bibitem{Lv2018}
S.~Lv, H.~Lin, H.~Lian, and J.~Huang.
\newblock Oracle inequalities for sparse additive quantile regression in
  reproducing kernel hilbert space.
\newblock {\em Annals of Statistics}, \textbf{2}:781--813, 2018.

\bibitem{Mendelson2010}
S.~Mendelson and J.~Neeman.
\newblock Regularization in kernel learning.
\newblock {\em Annal of Statistics}, \textbf{38}:526--565, 2010.

\bibitem{RadchenkoH2010}
P.~Radchenko and G.~James.
\newblock Variable selection using adaptive nonlinear interaction structures in
  high dimensions.
\newblock {\em Journal of the American Statistical Association},
  \textbf{105}:1541--1553, 2010.

\bibitem{Ritchie2001}
M.~Ritchie, L.~Hahn, N.~Roodi, L.~Bailey, W.~Dupont, F.~Parl, and J.~Moore.
\newblock Multifactor-dimensionality reduction reveals high-order interactions
  among estrogen-metabolism genes in sporadic breast cancer.
\newblock {\em The American Journal of Human Genetics}, \textbf{69}:138--147,
  2001.

\bibitem{Rosasco2013}
L.~Rosasco, S.~Villa, S.~Mosci, M.~Santoro, and A.~Verri.
\newblock Nonparametric sparsity and regularization.
\newblock {\em Journal of Machine Learning Research}, \textbf{14}:1665--1714,
  2013.

\bibitem{Shao2012}
J.~Shao and X.~Deng.
\newblock Estimation in high-dimensional linear models with deterministic
  design matrices.
\newblock {\em Annals of Statistics}, \textbf{40}:1821--1831, 2012.

\bibitem{She2018}
Y.~She, Z.~Wang, and H.~Jiang.
\newblock Group regularized estimation under structural hierarchy.
\newblock {\em Journal of the American Statistical Association}, in press,
  2018.

\bibitem{Shen2012}
X.~Shen, W.~Pan, and Y.~Zhu.
\newblock Likelihood-based selection and sharp parameter estimation.
\newblock {\em Journal of the American Statistical Association},
  \textbf{107}:223--232, 2012.

\bibitem{Shen2013}
X.~Shen, W.~Pan, Y.~Zhu, and Z.~Zhou.
\newblock On constrained and regularized high-dimensional regression.
\newblock {\em Annals of the Institute of Statistical Mathematics},
  \textbf{65}:807--832, 2013.

\bibitem{Shively1999}
T.~Shively, R.~Kohn, and S.~Wood.
\newblock Variable selection and function estimation in additive non-parametric
  regression using a data-based prior.
\newblock {\em Journal of the American Statistical Association},
  \textbf{94}:777--794, 1999.

\bibitem{Smale2005}
S.~Smale and D.~Zhou.
\newblock Shannon sampling ii: connections to learning theory.
\newblock {\em Applied and Computational Harmonic Analysis},
  \textbf{19}:285--302, 2005.

\bibitem{Smale2007}
S.~Smale and D.~Zhou.
\newblock Learning theory estimates via integral operators and their
  approximations.
\newblock {\em Constructive Approximation}, \textbf{26}:153--172, 2007.

\bibitem{Stafabski2014}
L.~Stefanski, Y.~Wu, and K.~White.
\newblock Variable selection in nonparametric classification via measurement
  error model selection likelihoods.
\newblock {\em Journal of the American Statistical Association},
  \textbf{109}:574--589, 2014.

\bibitem{Steinwart2008}
I.~Steinwart and A.~Christmann.
\newblock {\em Support Vector Machine}.
\newblock Springer, 2008.

\bibitem{SunWW2013}
W.~Sun, J.~Wang, and Y.~Fang.
\newblock Consistent selection of tuning parameters via variable selection
  stability.
\newblock {\em Journal of Machine Learning Research}, \textbf{14}:3419--3440,
  2013.

\bibitem{Szekely2007}
G.~Szekely, M.~Rizzo, and N.~Bakirov.
\newblock Measuring and testing dependence by correlation of distances.
\newblock {\em Annals of Statistics}, \textbf{35}:2769--2794, 2007.

\bibitem{Tibshirani1996}
R.~Tibshirani.
\newblock Regression shrinkage and selection via the lasso.
\newblock {\em Journal of the Royal Statistical Society, Series B},
  \textbf{58}:267--288, 1996.

\bibitem{Wahba1998}
G.~Wahba.
\newblock Support vector machines, reproducing kernel hilbert spaces, and
  randomized gacv.
\newblock {\em Advances in kernel methods: support vector learning}, pages
  69--88, MIT Press, {1998}.

\bibitem{Wang2009}
H.~Wang.
\newblock Forward regression for ultra-high dimensional variable screening.
\newblock {\em Journal of the American Statistical Association},
  \textbf{104}:1512--1524, 2009.

\bibitem{WangX2016}
X.~Wang and C.~Leng.
\newblock High dimensional ordinary least squares projection for screening
  variables.
\newblock {\em Journal of the Royal Statistical Society, Series B},
  \textbf{78}:589--611, 2016.

\bibitem{WuY2015}
Y.~Wu and L.~Stefanski.
\newblock Automatic structure recovery for additive models.
\newblock {\em Biometrika}, \textbf{102}:381--395, 2015.

\bibitem{YangL2016}
L.~Yang, S.~Lv, and J.~Wang.
\newblock Model-free variable selection in reproducing kernel hilbert space.
\newblock {\em Journal of Machine Learning Research}, \textbf{17}:1--24, 2016.

\bibitem{YangY2017}
Y.~Yang, M.~Pilanci, and M.~Wainwright.
\newblock Randomized sketches for kernels: fast and optimal nonparametic
  regression.
\newblock {\em Annals of Statistics}, \textbf{45}:991--1023, 2017.

\bibitem{ZhangC2010}
C.~Zhang.
\newblock Nearly unbiased variable selection under minimax concave penalty.
\newblock {\em Annals of Statistics}, \textbf{38}:894--942, 2010.

\bibitem{ZhangC2016}
C.~Zhang, Y.~Liu, and Y.~Wu.
\newblock On quantile regression in reproducing kernel hilbert spaces with data
  sparsity constraint.
\newblock {\em Journal of Machine Learning Research}, \textbf{17}:1--45, 2016.

\bibitem{ZhouD2007}
D.~Zhou.
\newblock Derivative reproducing properties for kernel methods in learning
  theory.
\newblock {\em Journal of Computational and Applied Mathematics},
  \textbf{220}:456--463, 2007.

\bibitem{ZouH2006}
H.~Zou.
\newblock The adaptive lasso and its oracle properties.
\newblock {\em Journal of the American Statistical Association},
  \textbf{101}:1418--1429, 2006.

\end{thebibliography}
		\bibliographystyle{plain}{}

    \begin{table}[!h]
    		\caption{{The averaged signal-to-noise ratio (SNR) of  the simulated examples under different scenarios.}}
    	\centering
    	\medskip
    	\label{tab:sim0}
    	\begin{tabular}{c|cccc}
    		\hline
    		$(n, \eta)$  & $(400, 0)$ & $(400,1)$ & $(500, 0)$ & $(500, 1)$ \\
    		\hline
    		Example 1  & 5.00  & 3.87  & 5.06& 3.87\\
    		\hline
    		Example 2 & 3.58 & 4.23  & 3.55 & 4.20\\
    		\hline
    	\end{tabular}
    \end{table}
    	
    \begin{table}[!h]
    	\caption{The averaged performance measures of various methods in Example 1.}
    	
    	\label{tab:sim1}
    	\begin{center}
    		\begin{tabular}{cc|cccccc}
    			\hline
    			$(n,p,\eta)$ & Method & Size & TP & FP & C & U & O \\
    			\hline
    			(400,500,0) & GM  & 5.00  & 5.00  & 0.00 & 50 & 0  &  0 \\
    			& QaSIS-t  & 4.28  & 4.28 & 0.00 & 22 & 28 & 0 \\
    			& DC-t  &  4.80  & 4.80  & 0.00  & 40 & 10 & 0  \\
    			\hline
    			(400,1000,0) & GM  &  4.98   & 4.98  & 0.00  & 49 & 1 & 0 \\
    			& QaSIS-t  & 4.32    & 4.32 & 0.00 & 21 & 29 & 0  \\
    			& DC-t  &  4.78   & 4.78   & 0.00 & 39 & 11 &  0 \\
    			\hline
    			(500,10000,0) & GM  & 5.00   & 5.00 & 0.00 & 50 & 0 & 0  \\
    			& QaSIS-t  & 4.28   & 4.28 & 0.00 & 24 & 26 & 0 \\
    			& DC-t  &  4.68  & 4.68 & 0.00 & 36 & 0 & 14 \\
    			\hline
    			(500,50000,0) & GM  &  5.06 & 4.98 & 0.08 & 45 & 1 & 4 \\
    			& QaSIS-t  &  4.08  & 4.08  & 0.00 & 18 & 32 & 0  \\
    			& DC-t  &  4.48   & 4.48  &  0.00& 28 & 22 & 0  \\
    			\hline
    			(500, 100000,0) & GM &  5.18   &  5.00 & 0.18 & 43 & 0  & 7    \\
    			& QaSIS-t  &   3.98  & 3.98  & 0.00  & 8 & 42 & 0 \\
    			& DC-t & 4.52  &  4.52  & 0.00 & 28  &  22& 0 \\
    			\hline
    			(400,500,1) & GM  & 4.98   & 4.98  & 0.00 & 49 & 1 & 0 \\
    			& QaSIS-t  &   2.80  & 2.72 & 0.08 & 0 & 50 & 0 \\
    			& DC-t&   2.94  & 2.94 & 0.00  & 0 & 50 & 0  \\
    			\hline
    			(400,1000,1) & GM  &  4.96  & 4.96  & 0.00 & 48 & 2 &  0 \\
    			& QaSIS-t &  2.34   & 2.26 & 0.08  & 0 & 50 & 0 \\            	       
    			& DC-t  &   2.96  & 2.96 &  0.00 & 0 & 50 & 0  \\
    			\hline
    			(500,10000,1) & GM  & 4.94  & 4.94 & 0.00 & 47 & 3 & 0  \\
    			& QaSIS-t  & 2.38    & 2.28 & 0.10 & 0 & 50 & 0 \\
    			& DC-t  &  3.08   & 3.08 & 0.00 & 0 & 50 & 0 \\
    			\hline
    			(500,50000,1) & GM  & 4.96  & 4.92 & 0.04  & 44 & 4 & 2  \\
    			& QaSIS-t  &  2.42   & 2.36 & 0.08 & 0 & 50 & 0  \\
    			& DC-t & 2.94    & 2.94 & 0.00 & 0 & 50 & 0  \\
    			\hline
    			(500, 100000, 1) & GM &  4.94  & 4.92  & 0.02 & 46 & 3 & 1  \\
    			& QaSIS-t  & 10.26    &  2.46 &  7.80 & 0 & 50 & 0  \\
    			& DC-t & 3.12   & 3.12  & 0.00  & 0  & 50 & 0   \\
    			\hline
    		\end{tabular}
    		
    	\end{center}
    \end{table}
    \begin{table}[!h]
    	\caption{The averaged performance measures of various methods in Example 2.}
    	\label{tab:sim2}
    	\begin{center}
    		\begin{tabular}{cc|cccccc}
    			\hline
    			$(n,p,\eta)$ & Method & Size & TP & FP & C & U & O  \\
    			\hline
    			(400,500,0) & GM  &  5.00   & 5.00 & 0.00 & 50 & 0 & 0\\
    			& QaSIS-t  & 4.26 & 4.26 & 0.00 & 22 &28  & 0  \\
    			& DC-t  & 4.92 & 4.92 & 0.00 & 48 & 2 & 0  \\
    			\hline
    			(400,1000,0) & GM  & 5.14  & 5.00    & 0.14 & 44 & 0 & 6  \\
    			& QaSIS-t  & 4.04 & 4.04 & 0.00 & 20 & 30 & 0 \\
    			& DC-t  & 4.96 & 4.96 & 0.00 & 48 & 2 & 0\\
    			\hline
    			(500,10000,0) & GM  & 5.10    & 5.00 & 0.10 & 45  & 0  & 5\\
    			& QaSIS-t &  3.82 & 3.82 & 0.00 & 13 & 37 & 0\\
    			& DC-t  & 4.92 & 4.92 & 0.00& 46 & 4 & 0 \\
    			\hline
    			(500,50000,0) & GM &  5.40   & 5.00 & 0.40 & 37 &  0 & 13\\
    			& QaSIS-t  & 3.04 & 3.04 & 0.00 & 8 & 42 & 0\\
    			& DC-t  &4.66 & 4.66 & 0.00 & 38 & 12 & 0 \\
    			\hline
    			(500, 100000,0) & GM   &   5.32  & 5.00  & 0.32   & 41 & 0 & 9 \\
    			& QaSIS-t  & 3.02    & 3.02  & 0.00  & 5 & 45 & 0 \\
    			& DC-t & 4.66   & 4.66   & 0.00 & 34  & 16 &  0 \\
    			\hline
    			(400,500,1) & GM & 5.00     & 4.98 & 0.02 & 48 & 1 & 1\\
    			& QaSIS-t  &  5.78   & 2.90 & 2.88 & 3 & 38  & 9\\
    			& DC-t &  31.30   & 4.00 & 27.30 & 1 & 0 & 49\\
    			\hline
    			(400,1000,1) & GM  & 5.10    &  5.00 & 0.10 & 45 & 0 & 5 \\
    			& QaSIS-t  &  7.78   & 2.22 & 5.56 & 1 & 42 & 7 \\
    			& DC-t  & 38.74    & 5.00 & 33.74 & 2 & 0 & 48 \\
    			\hline
    			(500,10000,1) & GM &  5.10   & 4.96 & 0.14 & 42  & 2 & 6 \\
    			& QaSIS-t  & 12.94    & 2.08 & 10.86 & 0 & 45 & 5\\
    			& DC-t  &  74.98   & 5.00 & 69.98 & 0 & 0 & 50 \\
    			\hline
    			(500,50000,1) & GM  &  5.16   & 4.98 & 0.18 &  43 & 1 & 6 \\
    			& QaSIS-t &  32.52   & 2.08 & 30.44 & 0 & 42 & 8 \\
    			& DC-t  &   79.62  & 5.00 & 74.62 & 0 & 1 & 49 \\	
    			\hline	
    			(500, 100000,1) & GM  &   5.10  & 4.96  &  0.14 & 44 & 2 & 4 \\
    			& QaSIS-t  &   42.32  & 2.54  &  39.78 & 0 & 44 & 6 \\
    			& DC-t &  79.94   &  4.88  & 75.06 & 0  & 6 & 44  \\
    			\hline		
    		\end{tabular}
    	\end{center}
    \end{table}

\begin{figure}[!h]
	\caption{  The scatter plots of the response against a number of selected variables by GM in the supermarket dataset. The solid lines are the fitted curve by local smoothing, and the dashed lines are the fitted mean plus or minus one standard deviation.	}
	\label{fig:20}
	\centering
	\begin{subfigure}[b]{0.45\textwidth}
		\includegraphics[width=\textwidth]{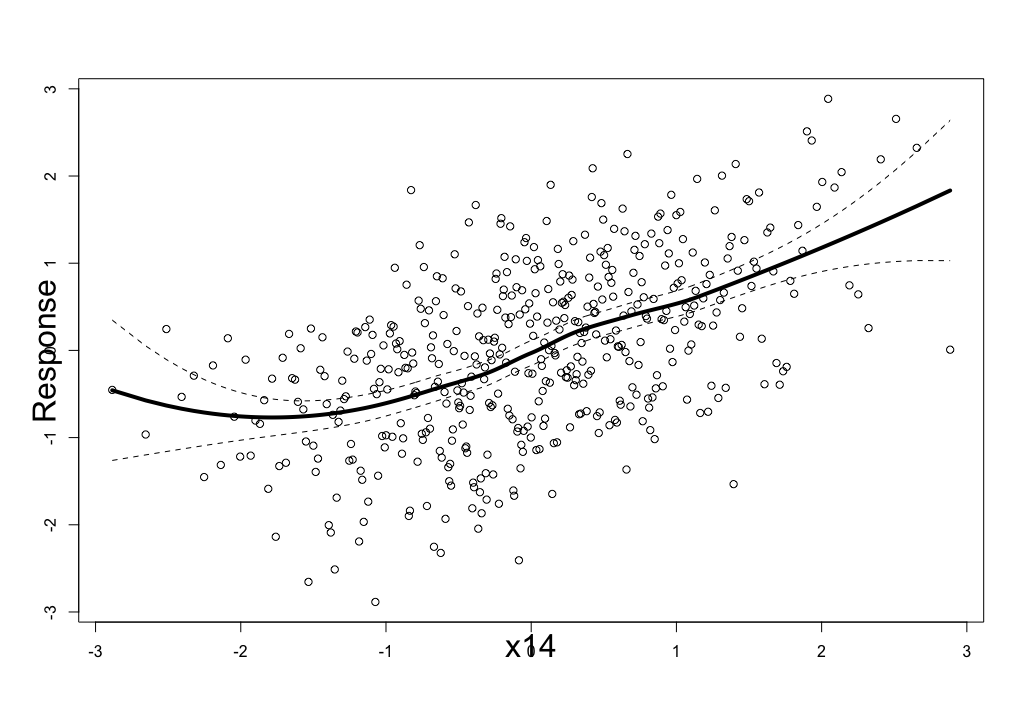}
		\caption{}
	\end{subfigure}
	\begin{subfigure}[b]{0.45\textwidth}
		\includegraphics[width=\textwidth]{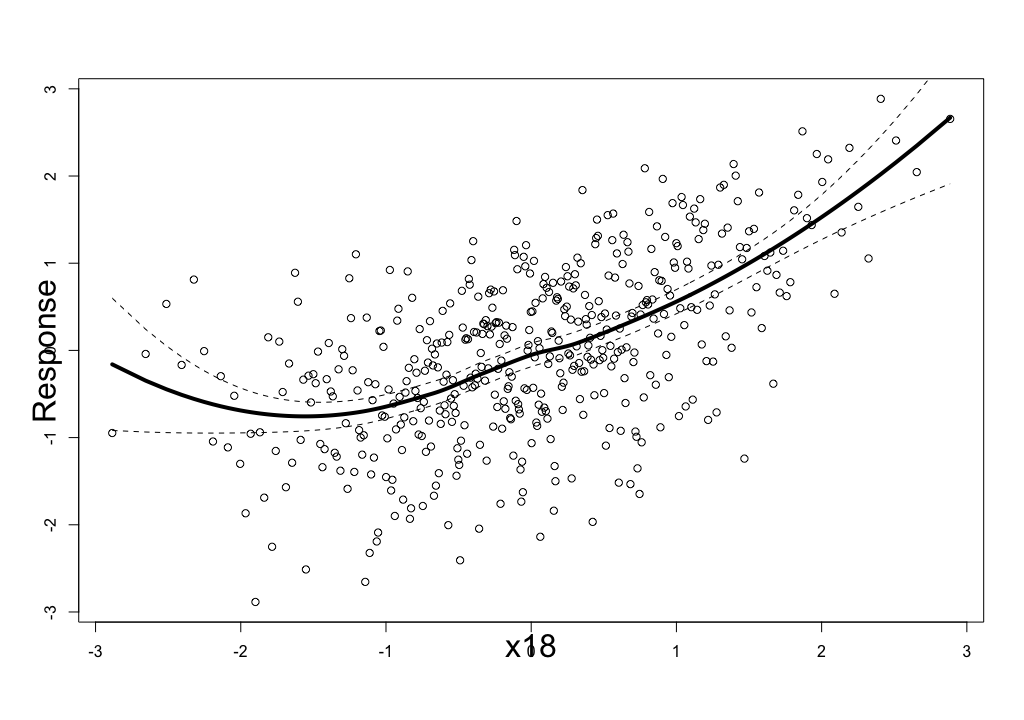}
		\caption{}
	\end{subfigure}
	\begin{subfigure}[b]{0.45\textwidth}
		\includegraphics[width=\textwidth]{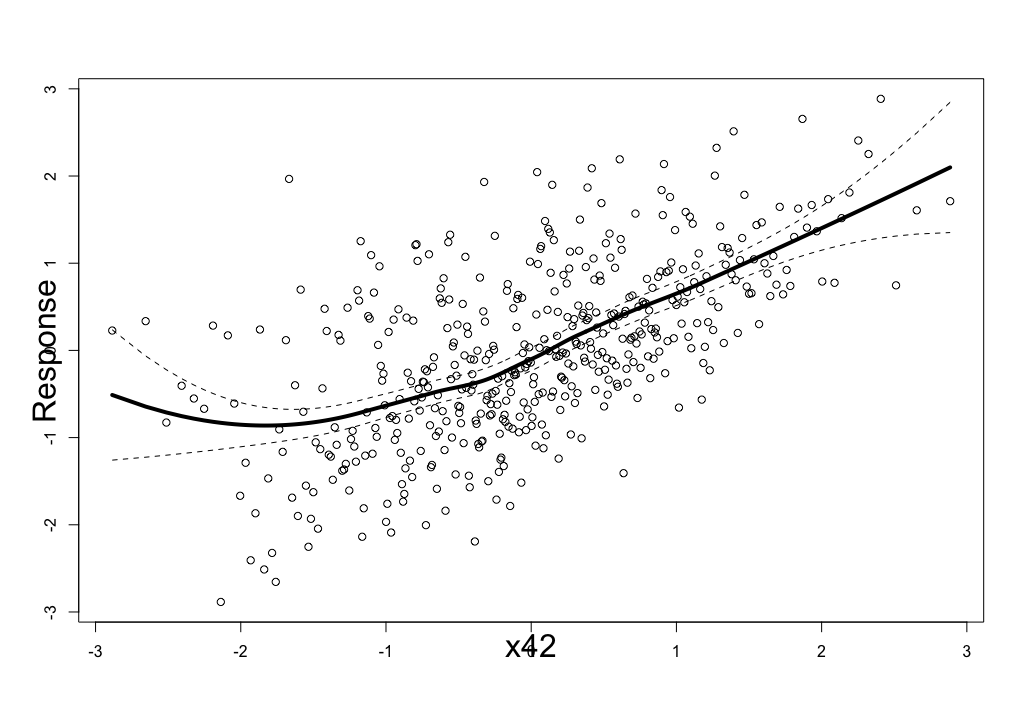}
		\caption{}
	\end{subfigure}
	\begin{subfigure}[b]{0.45\textwidth}
		\includegraphics[width=\textwidth]{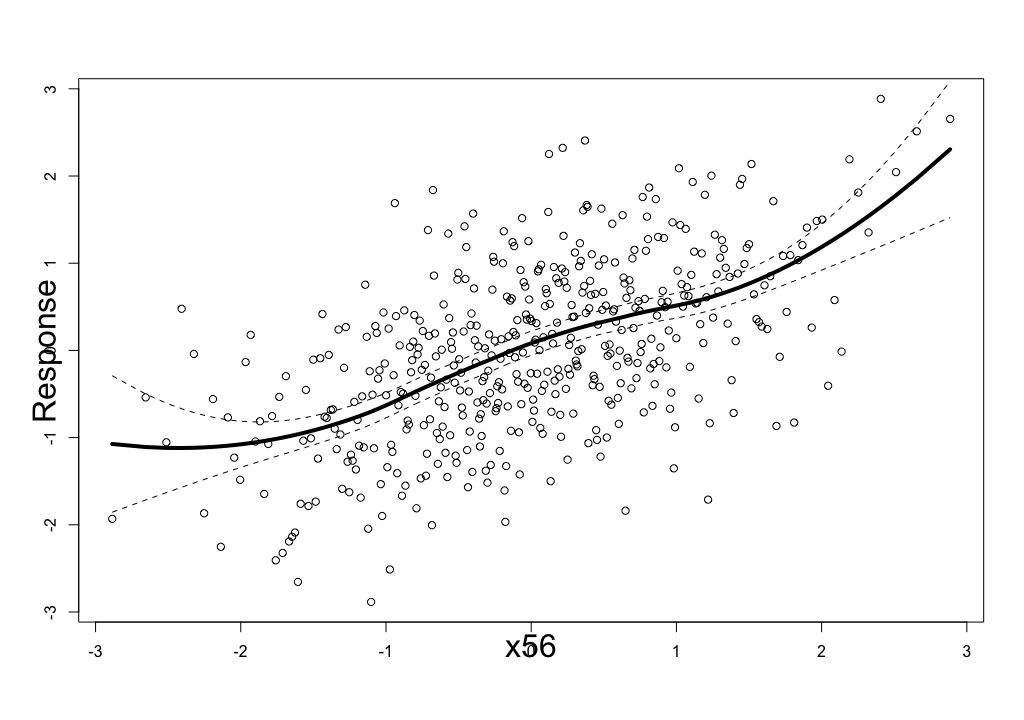}
		\caption{}
	\end{subfigure}
	\begin{subfigure}[b]{0.45\textwidth}
		\includegraphics[width=\textwidth]{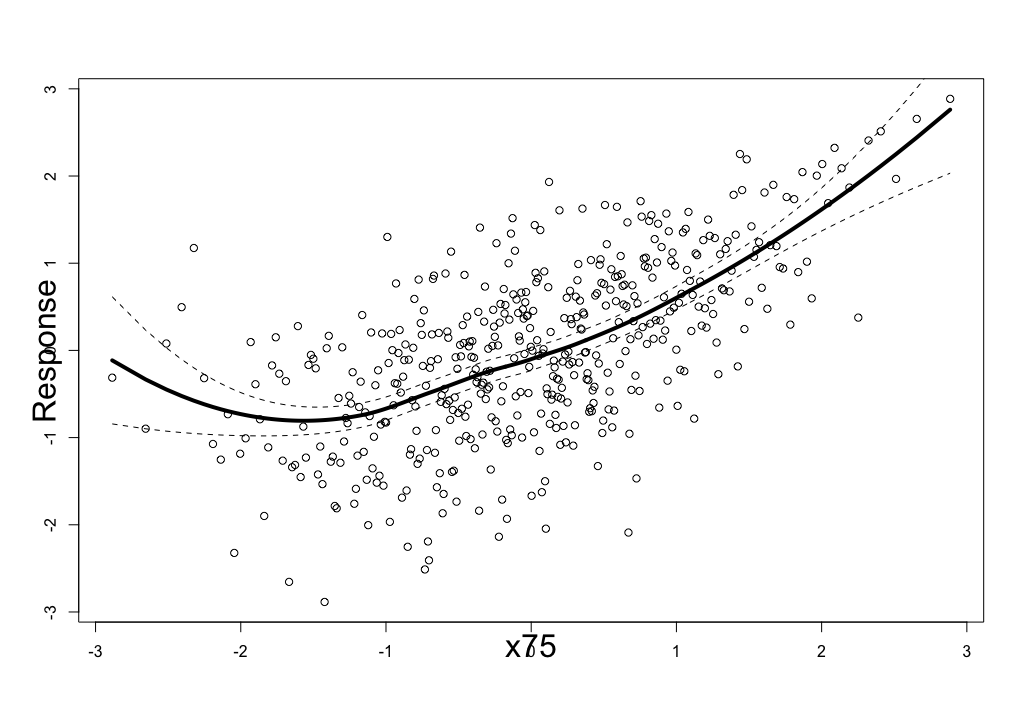}
		\caption{}
	\end{subfigure}
\end{figure}

\begin{table}[!h]
	\centering
	\caption{The number of selected variables as well as the corresponding averaged prediction errors by various methods in the supermarket dataset.}
	\label{tab:sim3}
	\begin{tabular}{ccccc}
		Dataset & Method &  Size & Testing error (Std) & Out of sample $R^2$ \\
		\hline
		& GM & 10 & 0.1369 (0.0005)  &   0.8631\\
		& QaSIS-t & 7 & 0.1674   (0.0006) & 0.8326\\
		& DC-t  & 7  &  0.1713(0.0006) & 0.8287\\
		&  SCAD     &  59  &  0.1872 (0.0006) & 0.8128\\
		&    MCP  &  28  &  0.2040 (0.0006)& 0.7960\\
		\hline
	\end{tabular}
\end{table}

\end{document}